\PassOptionsToPackage{numbers,sort&compress}{natbib}

\documentclass{article}

\usepackage[final]{corl_2026} 
\usepackage{xcolor}
\definecolor{mydarkblue}{rgb}{0,0.08,0.45} 
\hypersetup{
    colorlinks=true,
    linkcolor=mydarkblue,
    citecolor=mydarkblue,
    filecolor=mydarkblue,
    urlcolor=mydarkblue
}
\usepackage[T1]{fontenc}
\usepackage{parskip}
\usepackage{amsfonts}
\usepackage{dsfont}
\usepackage{amsmath}
\usepackage{amssymb}
\usepackage{anyfontsize}
\usepackage{bm}
\usepackage{enumitem}
\usepackage[utf8]{inputenc}
\newcommand{\anynorm}[1]{\left\lVert#1\right\rVert}
\usepackage{svg}
\usepackage{float}
\usepackage{tikz}
\usetikzlibrary{arrows.meta,positioning, calc, shadows, shapes.geometric}
\usepackage{caption}
\usepackage{subcaption}
\usepackage{graphicx}
\usepackage{pgfplots}
\pgfplotsset{compat=newest}
\usepackage{wrapfig}
\usepackage{cite}
\setcitestyle{numbers,sort&compress}

\usepackage{hyperref}
\usepackage[nameinlink]{cleveref}
\usepackage{booktabs}
\usepackage{siunitx}
\usepackage{tabularray}
\UseTblrLibrary{booktabs, siunitx}

\usepackage{algorithm}
\usepackage{algpseudocode}

\usepackage{xspace}
\newcommand{\SE}{\ensuremath{\text{SE}(3)}\xspace}
\newcommand{\SO}{\ensuremath{\text{SO}(3)}\xspace}

\title{RAM: Reachability Across Morphologies}

\author{
  Tim Walter\\
  Department of Computer Engineering\\
  Technical University of Munich\\
  Germany\\
  \texttt{tim.walter@tum.de} \\
  \And
  Xinyu Chen\\
  German Electron Synchrotron\\
  University of Hamburg\\
  Germany\\
  \texttt{xinyu.chen@desy.de}
  \And
  Jonathan K\"ulz\\
  Department of Computer Engineering\\
  Technical University of Munich\\
  Germany\\
  \texttt{jonathan.kuelz@tum.de} \\
  \And
  Matthias Althoff\\
  Department of Computer Engineering\\
  Technical University of Munich\\
  Germany\\
  \texttt{matthias.althoff@tum.de} \\
}

\begin{document}

\maketitle

\vspace{-1cm}
\begin{figure}[ht]
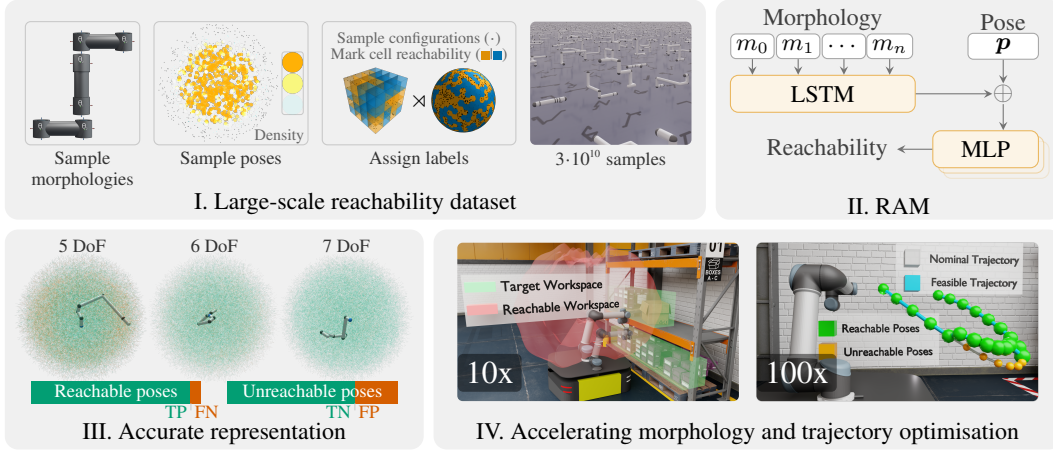

    \centering

    \definecolor{box}{RGB}{240,240,240}
    \definecolor{block}{RGB}{180,200,230}
    \definecolor{cardborder}{RGB}{200,200,200}
    \definecolor{SNSOrange}{RGB}{1., 115., 178.}
    \definecolor{SNSBlue}{RGB}{222., 143.,   5.}
    \definecolor{SNSGreen}{RGB}{2., 158., 115.}
    \definecolor{SNSRed}{RGB}{213.,  94.,   0.}

    \newcommand{\cbox}[1]{\tikz[baseline=-0.5ex]\node[fill=#1, minimum size=1.25ex, inner sep=0pt] {};}

    \newlength{\herounit}
    \setlength{\herounit}{\linewidth}
    \divide\herounit by 12

    \begin{tikzpicture}[
        x=\herounit, y=\herounit,   
        hero box/.style={
            rectangle, rounded corners=5pt, fill=box,
            minimum height=2.5\herounit, inner sep=0pt, align=center
        },
        card/.style={
            rectangle, draw=cardborder, fill=box, rounded corners=2pt,
            minimum height=1.4\herounit, minimum width=1.1\herounit, thin,
            inner sep=2pt, align=center
        },
        desc/.style n args={2}{
            label={[anchor=south, align=center, inner sep=0pt, yshift=2pt, font=\small]south:{#1. #2}}
        },
        card label/.style={
            font=\scriptsize, text=black!80, inner sep=1pt, align=center
        },
        card inner label/.style={
            font=\tiny, text=black!60, inner sep=1pt, align=center
        },
        block style/.style={
            rectangle, draw=black!70, fill=block, rounded corners=2pt,
            inner sep=1pt, align=center, font=\small, text=black!80
        },
        arrow style/.style={->, >=latex, thin},
    ]
        \pgfmathsetmacro{\thirdwidth}{(12 - 2*0.1)/3}
        \pgfmathsetmacro{\twothirdwidth}{12 - \thirdwidth - 0.1}
        \pgfmathsetmacro{\boxthreewidth}{0.4*(12 - 0.1)}
        \pgfmathsetmacro{\boxfourwidth}{0.6*(12 - 0.1)}

        \node[hero box, minimum width=\twothirdwidth\herounit,
              desc={I}{Large-scale reachability dataset}] (box1) {};
        \node[hero box, minimum width=\thirdwidth\herounit,
              desc={II}{RAM},
              right=0.1\herounit of box1] (box2) {};
        \node[hero box, minimum width=\boxthreewidth\herounit,
              desc={III}{Accurate representation},
              anchor=north west] (box3) at ($(box1.south west)+(0,-0.1\herounit)$) {};
        \node[hero box, minimum width=\boxfourwidth\herounit,
              desc={IV}{Accelerating morphology and trajectory optimisation},
              right=0.1\herounit of box3] (box4) {};

        \input{floats/source/hero/hero_box1.tex}
        \input{floats/source/hero/hero_box2.tex}
        \input{floats/source/hero/hero_box3.tex}
        \input{floats/source/hero/hero_box4.tex}

    \end{tikzpicture}
    \caption{We present RAM, a surrogate model for robot reachability. Trained on a diverse dataset of thirty billion samples (I), the morphology-conditioned implicit neural representation (II) generalises accurately to unseen morphologies (III). Due to its differentiability and efficient inference, RAM can substantially accelerate downstream robotics tasks by replacing inverse kinematics (IV).}
    \label{fig:hero}
\end{figure}

\begin{abstract}
    Many stages of the robotic lifecycle, from morphology synthesis to operation, rely fundamentally on the reachable workspace. 
    However, current methods for approximating workspaces are slow, imprecise, or tied to a single morphology.  We introduce Reachability Across Morphologies (RAM): a morphology-conditioned, implicit neural representation that acts as a fast, differentiable surrogate for pose reachability, generalising to unseen morphologies while inherently accounting for self-collisions. RAM is trained on a large-scale dataset of $3\cdot10^{10}$ samples generated solely from forward kinematics, which we publish alongside the model. Experiments show that RAM reaches 84\% balanced accuracy at nanosecond inference, 15 percentage points higher than a generative inverse kinematics baseline that also generalises across morphologies, while running three orders of magnitude faster. This speed carries over to downstream tasks, accelerating morphology and trajectory optimisation by one to two orders of magnitude.\\
    Website: \href{https://timwalter.github.io/ram}{https://timwalter.github.io/ram}.
\end{abstract}


\section{Introduction}
Tailoring robots to their operating environments through design optimisation, modular reconfiguration, or tool-use~\citep{KuelzCopilot2025,DaiReconfiguration2024, QintToolUse2023} commonly improves task performance, while reducing costs and complexity compared to general-purpose solutions~\citep{DerooEnergy2023, SchieleUnderactuated2022, HoffmanBiLevel2025, HaDesignopt2018, KuelzCopilot2025}. Despite these advantages, a major computational hurdle in both the design and operation of novel morphologies is computing their workspace, which requires solving inverse kinematics and checking for self-collisions~\citep{HartenbergDH1965, ShirafujiKinematicSynthesis2025}.
Inverse kinematics do not admit analytical solutions for arbitrary kinematic chains~\citep[Chapter 6.2]{LynchIK2017}, so numerical solvers are typically employed, yet these are slow, sensitive to initialisation, and converge to a single joint configuration rather than characterising the full solution set~\citep{PiperIK1969, LynchIK2017}. Moreover, exhaustive workspace evaluation requires querying millions of poses, making iterative design synthesis or reactive planning prohibitively slow.
\par 
Our key insight is that iterative tasks such as trajectory and morphology optimisation require only the existence of an inverse kinematics solution at each step, whereas the solution itself is only needed for the final result. While pre-computing workspaces has proven effective for individual robots~\citep{ZachariasCapabilty2007,ChirikjianConv1998, VahrenkampVoxel2012, HanSE2Conv2021, LeibrandtDiscretisation2023}, they must be recomputed for every morphology. We aim for a representation that generalises across different morphologies. To this end, we propose Reachability Across Morphologies (RAM): a fast, differentiable surrogate model for reachability. RAM is a morphology-conditioned, implicit neural representation~\citep{SitzmannINR2020} that encodes the reachable workspace through an occupancy network~\citep{MeschderOnet2019}. By shifting to a learned representation, we enable reachability evaluation in nanoseconds and ensure that the predicted workspace varies smoothly across the design space. Learning the intricate relationship between morphology and reachability requires large amounts of training data, which we efficiently generated without the computational overhead of inverse kinematics.
\par
We demonstrate that RAM represents diverse workspaces with high fidelity, outperforming the generalised inverse kinematics baseline~\citep{LimoyoGGIK2025} in both accuracy and inference latency. Furthermore, we show that our data generation produces reliable labels and scales to billions of samples. Finally, we demonstrate the practical utility of RAM by showing substantial acceleration in morphology and trajectory optimisation compared with numerical inverse kinematics.
\par 
In summary, our core contributions are:
\begin{itemize}
    \item \textbf{RAM}: a novel, morphology-conditioned implicit neural representation based on an occupancy network that acts as a fast, differentiable surrogate model for pose reachability.
    \item \textbf{Large-scale Reachability Dataset}: A public dataset of $3\cdot10^{10}$ samples across $3\cdot10^4$ unique robot morphologies generated solely from forward kinematics and self-collision queries, with pose sampling restricted to geometrically plausible regions to expedite learning.
    \item \textbf{Accelerating Morphology \& Trajectory Optimisation}: Our differentiable surrogate model enables morphology and trajectory optimisation at significantly lower computational costs compared to numerical inverse kinematics.
\end{itemize}

\section{Problem Statement}
Our objective is to learn the reachability map $\Psi:\mathcal{M} \times \SE \to \left\{0,1\right\}$. This map identifies whether a pose $\bm{P} \in \SE$ lies in the reachable workspace of a morphology $\bm{M} \in \mathcal{M}$. The reachable workspace is the set of all end-effector poses attained by the forward kinematics $f:\mathcal{M} \times \mathcal{T}^n \to \SE$ over all collision-free joint configurations $\bm{\theta} \in \mathcal{T}^n$, with the $n$-dimensional torus $\mathcal{T}^n$ and the number of joints $n$. After introducing $f^{-1}\left(\bm{M},\bm{P}\right)=\left\{\bm{\theta}\in\mathcal{T}^n\mid f\left(\bm{M},\bm{\theta}\right)=\bm{P}\right\}$ which returns the set of inverse kinematic solutions and $c\left(\bm{M}, \bm{\theta}\right)$ which holds if morphology $\bm{M}$ in configuration $\bm{\theta}$ is not in self-collision, we can define the reachability map as
\begin{equation}
    \Psi\left(\bm{M},\bm{P}\right) = \begin{cases}
        1 & \text{if} \quad \exists \bm{\theta} \in f^{-1}\left(\bm{M},\bm{P}\right): c\left(\bm{M}, \bm{\theta}\right)\\
        0 & \text{else}.
    \end{cases}
\end{equation}

\section{Related Work}
\textbf{Inverse Kinematics.}
Traditionally, reachability is determined by solving the inverse kinematics problem directly. Analytical solvers such as IKfast~\citep{DiankovIKFast2018} and the Efficient Analytical Inverse Kinematics toolbox (EAIK)~\citep{OstermeierEAIK2025} are reliable, fast, and yield the complete solution set $f^{-1}\left(\bm{M},\bm{P}\right)$ for morphologies with decomposable topologies or fewer than five degrees of freedom (DoF). Numerical solvers apply to arbitrary morphologies by iteratively minimising a pose residual  $r\left(\bm{P}, f(\bm{M},\bm{\theta})\right)$ over $\bm{\theta}$, typically with multiple random restarts to improve coverage of the solution space~\citep{YasutakeHJCDIK2025, BeesontracIK2015, SundaralingamCuRobo2023}. Learning-based models of inverse kinematics~\citep{OehsenLearnIK2020, AmesIKFlow2022, BensadounIKNet2022, LimoyoGGIK2025, KuelzCopilot2025} trade accuracy for speed, with generative approaches even predicting diverse solution sets. Yet, they still predict a joint configuration $\bm{\theta}$, which is a richer output than the reachability map $\Psi$ requires.
\par
\textbf{Reachability Maps.}
For a fixed morphology, pre-computing its workspace as a map reduces reachability assessment to a lookup. Traditional approaches range from extracting workspace boundaries through singularity analysis~\citep{AbdelMalekStratification2006} to compact geometric approximations such as ellipsoids~\citep{CavelliEllipsoid2025}. However, singularity analysis requires enumerating joint limit combinations that grow combinatorially with the number of joints, limiting applicability to five DoF, while geometric approximations sacrifice fidelity for compactness, motivating discretised alternatives. Voxel-based maps were first introduced by \citet{ZachariasCapabilty2007} to pre-calculate reachability over the entire workspace via inverse kinematics. Subsequent work~\citep{ChirikjianConv1998, VahrenkampVoxel2012, HanSE2Conv2021, LeibrandtDiscretisation2023} extended them from positions to full \SE poses and self-collision measures, and expedited the computation by convolving reachability maps of single links. Despite these advances, voxel-based methods remain fundamentally memory-intensive and tied to a single morphology.
\par
\textbf{Implicit Neural Representations.}
Implicit neural representations have emerged as a powerful alternative to discretisation by approximating continuous functions with neural networks~\citep{SitzmannINR2020}. In computer vision and graphics, representations such as neural signed distance functions~\citep{ParkDeepSDF2019}, neural radiance fields~\citep{MildenhallNeRF2020}, and occupancy networks~\citep{MeschderOnet2019} have demonstrated that neural networks operating on spatial coordinates can represent complex geometries and volumetric scenes with high fidelity at a fraction of the memory costs of voxels. This paradigm has recently transitioned to robotics~\citep{IrshadNeuralFieldSurvey2024}, where such representations are used to encode robot geometry~\citep{LiSDFGeometry2024} and configuration-space distance fields~\citep{LiCDF2024}. Prior work has learned surrogate reachability models for single morphologies~\citep{GiengerRandomforests2024, MurookaOnlyFK2025, SeungsuFlow2021}, with \citet{GiengerRandomforests2024} incorporating self-collisions and \citet{MurookaOnlyFK2025} training solely from forward kinematics. \citet{Xinyu2024} explore morphology-conditioning but remain limited to 3D positions and a small design space. Full \SE reachability encompasses both orientation and position, yielding a substantially larger non-Euclidean space. The resulting workspaces exhibit far greater topological variation than position-only workspaces, and no prior work provides a morphology-conditioned neural representation of the full \SE reachable workspace that inherently accounts for self-collisions.

\section{Reachability Across Morphologies (RAM)}
We treat workspace representation as a binary classification problem. Our data synthesis first samples
a morphology, determines its geometrically plausible poses, and assigns reachability labels. A reachability classifier is then trained to implicitly encode the workspace in its decision boundary, i.e., the set of poses at which the predicted reachability equals 0.5, separating poses classified as reachable from those classified as unreachable.

\subsection{Morphology Sampling} \label{ssec:morphology_sampling}
\begin{figure}
 \centering
\subcaptionbox{$\bm{m} = [0, 1, 0]$\label{fig:type_2}}[0.32\linewidth]{\centering\includegraphics[scale=0.1]{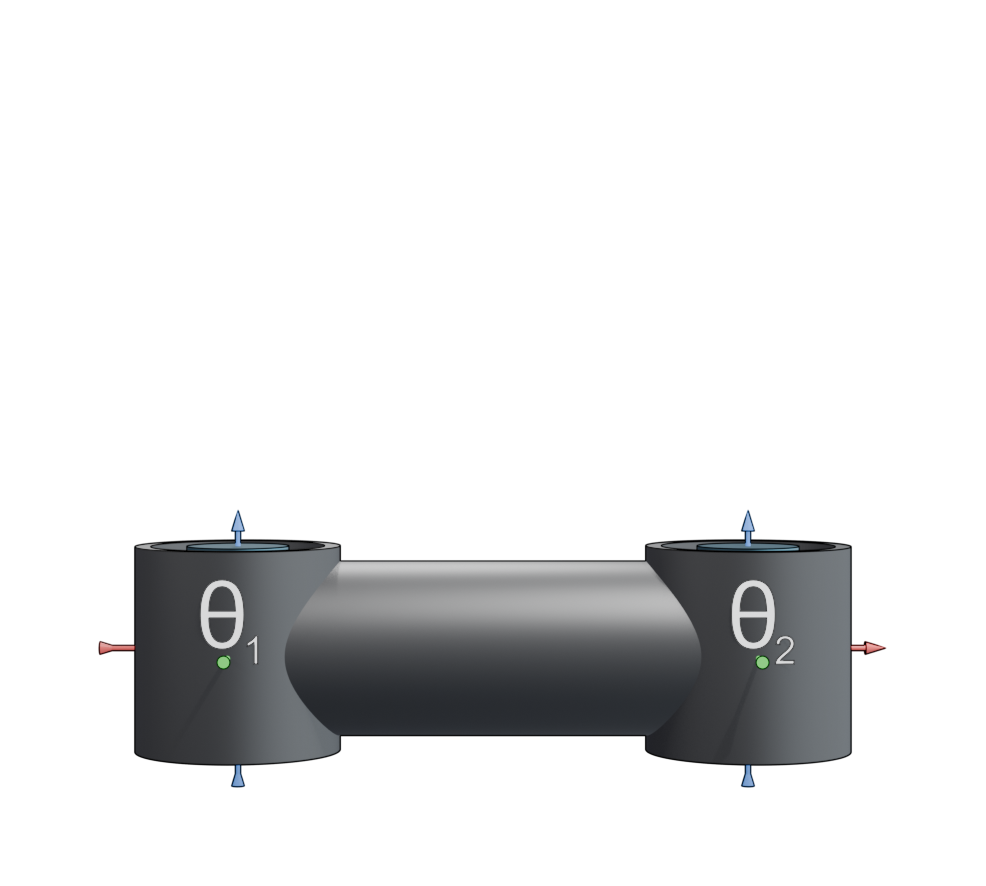}}\hfill
 \subcaptionbox{$\bm{m} = [0, 0, 1]$\label{fig:type_1}}[0.32\linewidth]{\centering\includegraphics[scale=0.1]{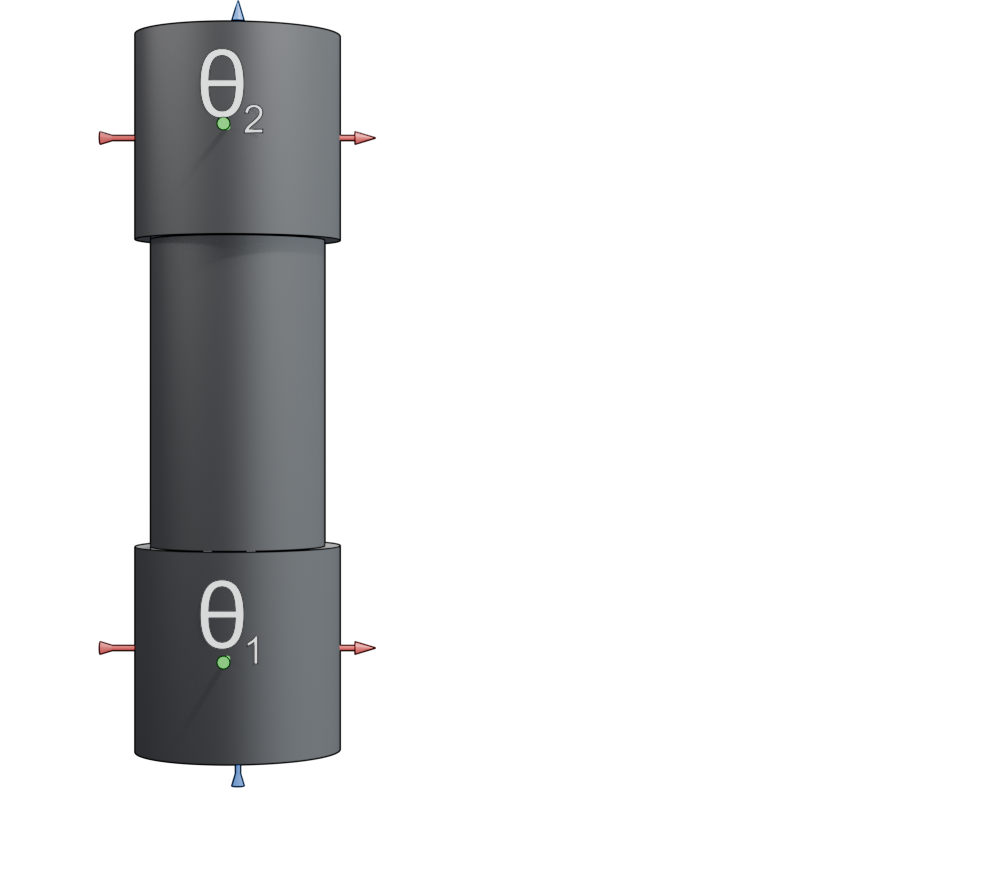}}\hfill
\subcaptionbox{$\bm{m} = [0, 1, 1]$\label{fig:type_0}}[0.32\linewidth]{\centering\includegraphics[scale=0.1]{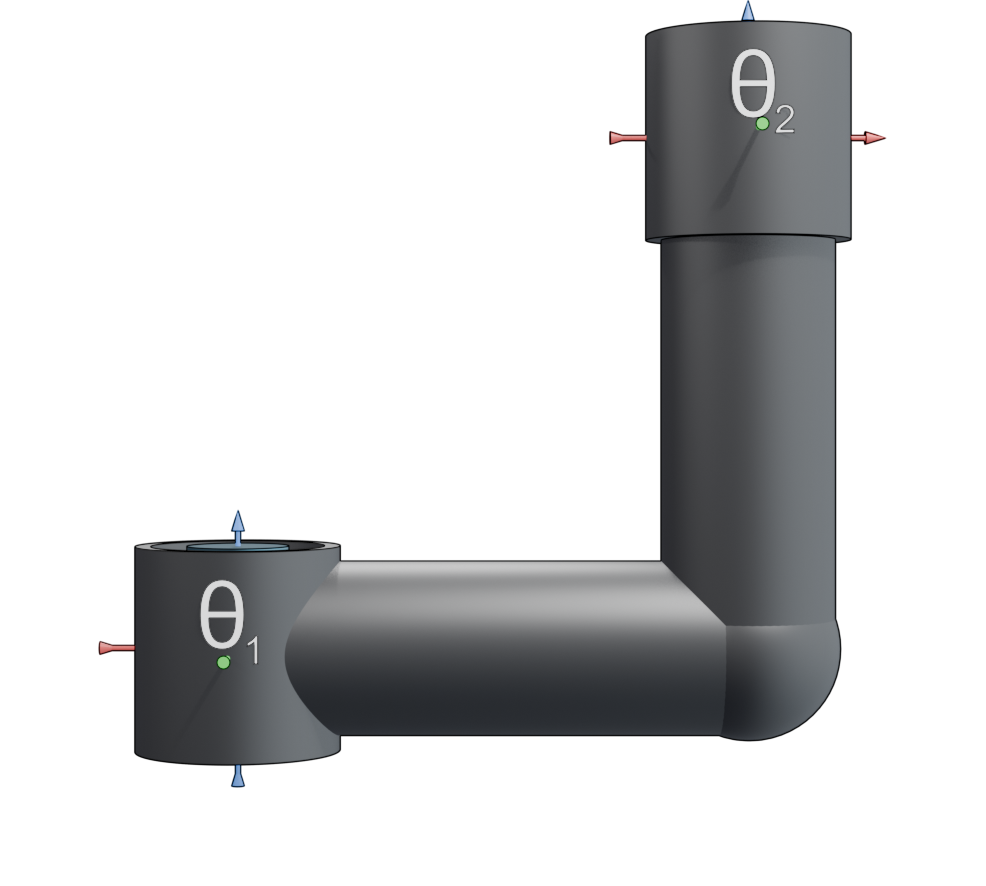}}

 \vspace{-1.45cm}
 
 \subcaptionbox{$\bm{m} = [\frac{\pi}{2}, 1, 0]$\label{fig:type_3}}[0.32\linewidth]{\centering\includegraphics[scale=0.1]{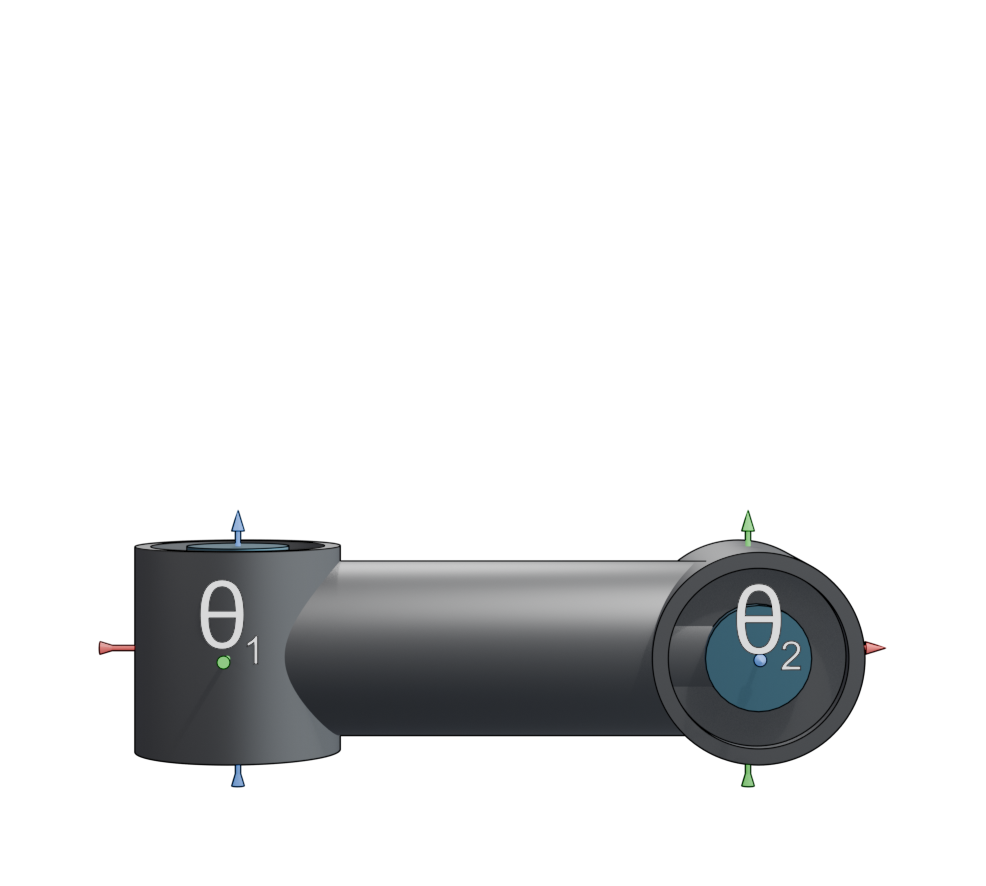}}\hfill
\subcaptionbox{$\bm{m} = [\frac{\pi}{2}, 0, 1]$\label{fig:type_4}}[0.32\linewidth]{\centering\includegraphics[scale=0.1]{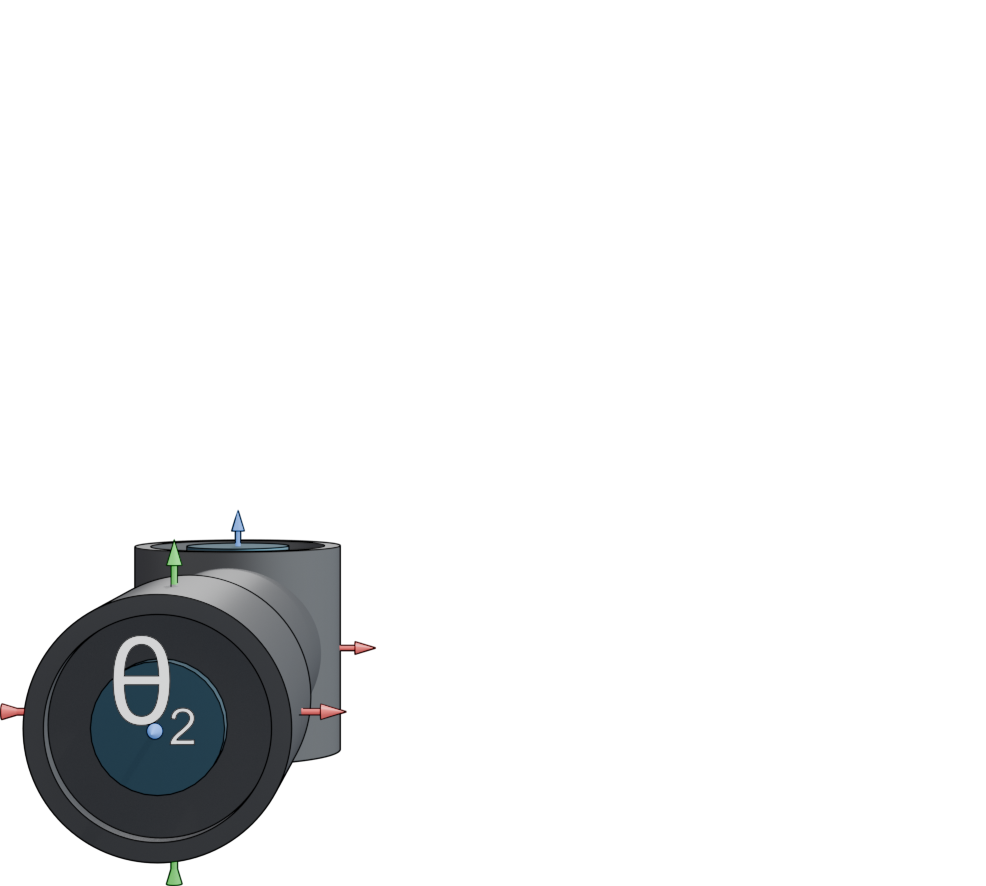}}\hfill
\subcaptionbox{$\bm{m} = [\frac{\pi}{2}, 1, 1]$\label{fig:type_5}}[0.32\linewidth]{\centering\includegraphics[scale=0.1]{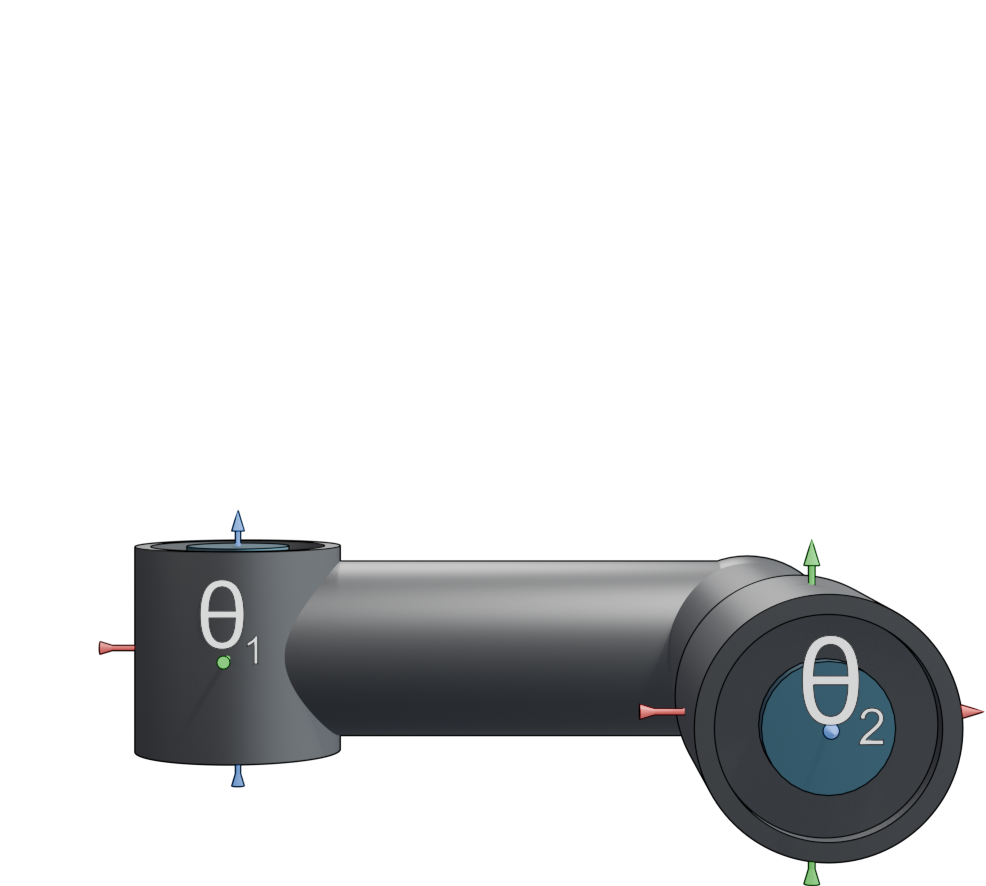}}
 \caption{Example link types with columns representing length-only, offset-only, and both. The bottom row (d-f) introduces a twist to the respective types in the top row (a-c). Joint angles are zero.}
 \label{fig:link_types}
\end{figure}

We consider serial kinematic chains of the form
\begin{equation*}
    \text{base}\rightarrow\text{joint}_{0} \rightarrow \text{link}_{1} \rightarrow \text{joint}_{1} \rightarrow\dots \rightarrow\text{joint}_{n-1} \rightarrow \text{end effector}
\end{equation*}
composed of $n\in\left\{5,6,7\right\}$ revolute joints covering the most practically relevant manipulator configurations that do not always admit analytical inverse kinematics. We restrict ourselves to morphologies that are:
\begin{enumerate}
    \item \textbf{Operable}: The morphology must admit at least one configuration free of self-collisions.
    \item \textbf{Non-degenerate}: We exclude degenerate designs, where degeneracy is defined as having fewer functional DoF $n_\text{func}$ than nominal joints $n_\text{func} < \min(6, n)$~\citep{ConkurRedundancy1997}.
    \item \textbf{Conventional}: Successive joint axes are either parallel or orthogonal, and no more than two joints are co-located.
\end{enumerate}
Modified Denavit-Hartenberg parameters~\citep{CraigMDH2013} encode our kinematics as the sequence $\bm{M} = \left(\bm{m}_0, \dots, \bm{m}_{n-1},\bm{m}_{\text{eef}}\right)$, where $\bm{m}_i = \begin{bmatrix} \alpha_{i-1}& a_{i-1}& d_i \end{bmatrix}$ encodes one link-joint pair via the link twist $\alpha_{i-1} \in \left\{-\frac{\pi}{2}, 0, \frac{\pi}{2} \right\}$, the link length $a_{i-1} \in \mathbb{R}$, and the link offset $d_i \in \mathbb{R}$. We collect all twists, lengths, and offsets into vectors $\bm{\alpha} \in \left\{-\frac{\pi}{2}, 0, \frac{\pi}{2}\right\}^n$, $\bm{a} \in \mathbb{R}^n$, $\bm{d} \in \mathbb{R}^n$ respectively. Since the workspace is topologically invariant to scale, we normalise without loss of generality to $\sum_i\sqrt{a_i^2+d_i^2}=1$. We assign a geometry to each pair by placing two capsules of radius $r$  with lengths $a_{i-1}$ and $d_i$ along the $x$- and $z$-axes of each link-joint frame, enabling efficient self-collision checks. \Cref{fig:link_types} shows example links.

\par
The morphological constraints create a complex feasible region, making direct uniform sampling impractical. Therefore, we employ hierarchical rejection sampling, detailed in \Cref{app:morph}, which exploits the dependence of later parameters on earlier ones to check constraints early and discard violating candidates from a large parallel batch. This enables efficient generation and uniform coverage of the valid morphological space, which is necessary to obtain unbiased training data.

\subsection{Pose Sampling} \label{ssec:pose_sampling}
We restrict pose sampling to geometrically plausible regions by excluding trivially unreachable positions outside the unit ball $\mathbb{R}^3\setminus\mathcal{B}^3$, a consequence of the unit-size normalisation of the robot. Moreover, most poses in $\mathcal{B}^3 \rtimes \SO$, where $\rtimes$ denotes the semi-direct product, are also unreachable: uniformly sampled positions tend to lie far from the base, since the volume of $\mathcal{B}^3$ concentrates near its boundary, where the robot must be nearly fully extended and can therefore only reach a highly restricted set of orientations. We therefore sample uniformly from the reduced pose space $\mathcal{B}^3\left(\bm{t}_0, r_{\text{mov}}\right)\rtimes\SO$, centred at the first joint position $\bm{t}_0$ with moveable length $r_\text{mov} = 1-\sqrt{a_0^2+d_0^2}-\sqrt{a_{\text{eef}}^2+d_{\text{eef}}^2}$, which is detailed in \Cref{app:pose_sampling}. Applying the end-effector transformation to these sampled poses yields geometrically plausible target poses. \hyperref[fig:hero]{Figure 1-I} depicts a typical sampling distribution. Morphologies with five DoF require special treatment, since their workspace is a five-dimensional submanifold embedded in six-dimensional \SE, which has measure zero. Consequently, the aforementioned sampling would yield a severe class imbalance towards unreachable poses. We therefore supplement the pose-sampled data with samples drawn directly from the configuration space via forward kinematics.

\subsection{Label Assignment}\label{ssec:label_assignment}
Rather than assessing each pose individually via inverse kinematics, we first approximate the reachable workspace and then determine reachability by checking whether a pose is contained within it. We employ the closed-world assumption~\citep{ReiterClosedworld1978}, i.e., that any fact not explicitly observed is assumed false, which in our context implies that any pose not reached during a sufficiently dense sampling of the configuration space is considered unreachable. To enable constant-time reachability queries, we discretise $\SE = \mathbb{R}^3 \rtimes \SO$ separately for $\mathbb{R}^3$ and \SO by exploiting its semi-direct product structure. While the translation space $\mathbb{R}^3$ is trivially split into cubic cells, \SO requires a more complex approach to reduce non-uniformity of the spatial discretisation. We employ the technique of \citet{LeibrandtDiscretisation2023}, which discretises \SO via iterative tessellation of the 600-cell~\citep[P.249]{Johnson600-Cell2018}, the 4D analogue of the icosahedron, and pre-computes a lookup table for constant-time orientation indexing. Given the discretisation, we approximate the workspace by sampling joint configurations, as detailed in \Cref{app:label_assignment}, evaluating the resulting forward kinematics, and discarding poses resulting in self-collision. Cells are marked as reachable if any valid pose is contained in them, which tends to over-approximate the true workspace, akin to a dilation of the workspace by up to one cell. Since a pose may admit multiple inverse kinematics solutions, a self-colliding configuration does not mark its cell as unreachable, as another solution may be collision-free. We label the poses sampled according to \Cref{ssec:pose_sampling} with the label of their respective cell. Once the workspace is approximated, labelling a new pose requires only a constant-time cell lookup, making dataset extension computationally negligible.

\subsection{Training \& Architecture} \label{ssec:training}
To ensure a continuous representation, we represent poses as $\bm{p} =\begin{bmatrix}\bm{t} & \bm{q}\end{bmatrix}^T\in \mathbb{R}^9$, where $\bm{t}\in\mathbb{R}^3$ denotes the translation vector and $\bm{q} \in \mathbb{R}^6$ is the continuous rotation representation of~\citet{ZhouContinuousrot2019}. Since $\bm{M}$ is a variable-length sequence, we encode it into a fixed-size latent embedding using a long short-term memory (LSTM)~\citep{HochreiterLSTM1997} module; given the short sequences considered, we opt for an LSTM over a transformer~\citep{vaswaniAttentionAllYou2017}, which achieves higher accuracy at lower computational overhead as \Cref{app:transformer} shows. This embedding is concatenated with $\bm{p}$ and passed through a multilayer perceptron (MLP), whose hidden layers use ReLU activations and whose final layer uses a sigmoid activation, yielding the reachability probability
\begin{equation}
    y=\text{RAM}\left(\bm{M}, \bm{p}\right)=\text{MLP}\left(\begin{bmatrix} \bm{p} \\\text{LSTM}\left(\bm{M}\right) \end{bmatrix}\right)\,.
\end{equation}
The training minimises the binary cross entropy loss $\mathcal{L}(l,y) = -\left(l\log \left(y\right)+(1-l)\log \left(1-y\right)\right)$, where $l\in\left\{0,1\right\}$ is the reachability label.

\section{Experiments}
We structured our experiments around three research questions:
\begin{description}[style=multiline, leftmargin=3em, font=\bfseries, nosep]
    \item[RQ1:] \textbf{Label fidelity.} How finely must \SE be discretised, and how many forward kinematics samples are required to achieve high-fidelity labels?
    \item[RQ2:] \textbf{Representation Accuracy.} How accurately and efficiently does RAM represent workspaces of unseen morphologies? 
    \item[RQ3:] \textbf{Practical Utility.} How does RAM influence the runtime and outcome of morphology and trajectory optimisation?
\end{description}
Across all experiments, we split morphologies equally among DoFs and poses equally among morphologies. We labelled the training set according to \Cref{ssec:label_assignment}, while we obtained the test and validation labels from a feasible variant of the reachability map $\Psi$
\begin{equation} \label{eq:exp_reachability}
    l = \begin{cases}
        1 & \text{if} \quad \exists \bm{\theta} \in \mathcal{Q}: \left( \anynorm{ f\left(\bm{M}, \bm{\theta}\right) - \bm{P}}_{\SE} < \epsilon \right)\wedge c\left(\bm{M}, \bm{\theta}\right)\\
        0 & \text{else}\,,
    \end{cases}
\end{equation}
which replaces exact reachability with a tolerance $\epsilon$ on the metric defined in  \Cref{app:preliminaries} and the full fibre $f^{-1}\left(\bm{M},\bm{P}\right)$ with the finite solution set $\mathcal{Q}$ from either EAIK~\citep{OstermeierEAIK2025} for analytically solvable kinematics or a collision-aware Levenberg-Marquardt solver with random restarts. We report all stochastic metrics as macro-averages over poses and morphologies, with 95\% confidence intervals via bootstrapping; where multiple training runs were conducted, we additionally average over seeds. The experiments were conducted on an Nvidia H100 GPU.

\subsection{Label Fidelity (RQ1)} \label{ssec:rq1}
Prior to generating training data, we investigated different discretisation granularities to ensure label fidelity and computational feasibility. The discretisation scheme of \Cref{ssec:label_assignment} only allows for a stepped modulation of granularity, since we desire parity between $\mathbb{R}^3$ and \SO and each subdivision of the 600-cell multiplies the number of \SO cells by $\approx2^3$~\citep{LeibrandtDiscretisation2023}. Since the tessellation of \SO is not perfectly uniform, we report the resulting cell spacing, i.e., the distance between the centres of neighbouring cells, as the interval between its smallest and largest value. We evaluated the workspace approximation at these granularities for $90$ morphologies and $9 \cdot 10^6$ poses. The approximation terminated either once the true positive rate exceeded 95\%, indicating convergence, or after a maximum runtime of 600 seconds. The implementation utilised accelerators~\citep{SchuckScipy2025} to sample joint configurations, perform forward kinematics, and label the resulting poses at roughly $5\cdot10^{7}$ configurations per second.
\par
\begin{table}
    \centering
    \caption{Effects of the discretisation granularity on workspace approximation.}
    \label{tab:discretisation_fidelity}
    \begin{tblr}{
            colspec = {l r r r},
            row{1} = {font=\bfseries},
        }
        \toprule
        Cell Spacing & \# Cells & Runtime (s) & Balanced Accuracy (\%) \\
        \midrule
        $\left[0.158, 0.163\right]$ & $231,840$ & $<1$ & \num{71(2:3)} \\\addlinespace
        $\left[0.080, 0.083\right]$ & $11,971,800$ & $1\,^{+0}_{-0}$ & \num{83(2:2)} \\\addlinespace
        $\left[0.040, 0.042\right]$ & $686,811,600$ & \num{24(4:4)} & \num{87(2:2)} \\\addlinespace
        $\left[0.021, 0.033\right]$ & $29,605,439,136$ & \num{394(47:46)} & \num{87(2:2)} \\\addlinespace
        \bottomrule
    \end{tblr}
\end{table}

\Cref{tab:discretisation_fidelity} shows that the balanced accuracy increased with finer granularity due to smaller cells, which reduced false positives as observable in the binary confusion matrix in \Cref{app:rq1}. However, at the finest granularity, not all approximations achieved 95\% true positives within the maximum runtime, resulting in more false negatives. Consequently, we used a cell spacing of $\left[0.040, 0.042\right]$ with a conservative runtime of 30s to approximate the workspace of each training morphology. The resulting labels have an expected balanced accuracy of 87\% and establish a practical ceiling on the accuracy achievable by RAM.

\subsection{Accuracy of RAM (RQ2)} \label{ssec:rq2}
Using our workspace approximation, we generated a training set of $3\cdot10^{10}$ samples from $3\cdot10^4$ distinct morphologies; see \Cref{app:dataset} for details. We trained RAM on ten different seeds with the hyperparameters reported in \Cref{app:hyperparameters}. Generative Graphical Inverse Kinematics (GGIK)~\citep{LimoyoGGIK2025} served as the baseline, since it is, to the best of our knowledge, the only published method generalising reachability prediction across morphologies in \SE. GGIK was trained using their out-of-distribution generalisation procedure~\citep[Sec. V.B]{LimoyoGGIK2025}, which required $4\cdot10^6$ distinct samples from our morphological space, generating 32 solutions per pose to yield $\mathcal{Q}$. Due to the long training times, we trained GGIK once. The validation and test sets consisted of $3\cdot10^{7}$ samples across $3\cdot10^2$ morphologies; due to the slow inference of GGIK, we evaluated it on 1\% of the samples per morphology. We selected the pose error threshold $\epsilon$ for GGIK by sweeping over candidate values and choosing the one that maximised balanced accuracy on the validation set.

\par
\begin{wrapfigure}{r}{0.3\textwidth}
    \vspace{-13em}
    \centering
    \subcaptionbox{Labels.\label{fig:slice_ground-truth}}{
        \begin{tikzpicture}
            \node[draw, black, line width=1pt, inner sep=0pt] at (0,0)
                {\includegraphics[width=0.28\textwidth]{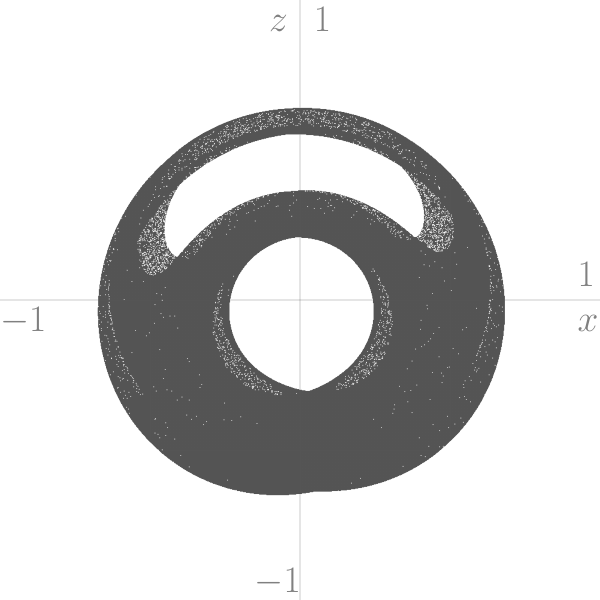}};
        \end{tikzpicture}
    }\\[0.5ex]
    \subcaptionbox{RAM.\label{fig:slice_ram}}{
        \begin{tikzpicture}
            \node[draw, black, line width=1pt, inner sep=0pt] at (0,0)
                {\includegraphics[width=0.28\textwidth]{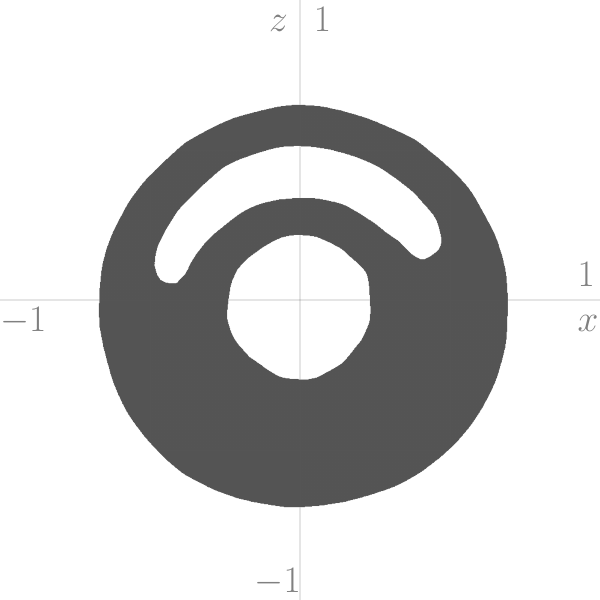}};
        \end{tikzpicture}
    }\\[0.5ex]
    \subcaptionbox{GGIK.\label{fig:slice_ggik}}{
        \begin{tikzpicture}
            \node[draw, black, line width=1pt, inner sep=0pt] at (0,0)
                {\includegraphics[width=0.28\textwidth]{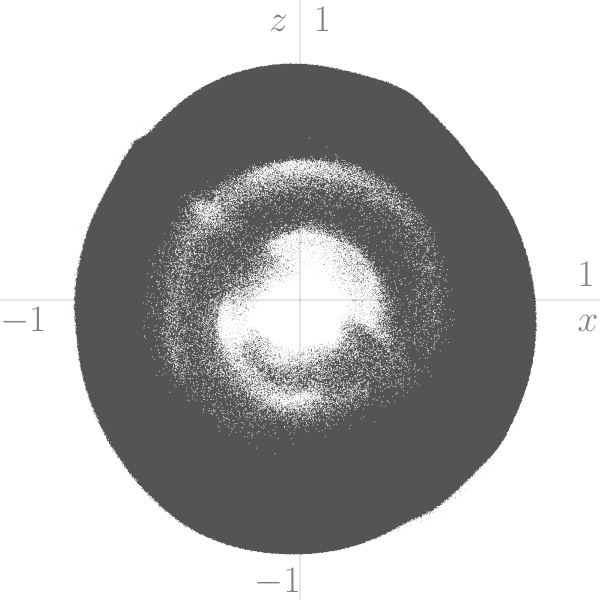}};
        \end{tikzpicture}
    }
    \vspace{-0.5em}
    \caption{Example of a work\-space slice with reachable poses indicated by dark pixels. Generated according to \Cref{app:slices}}
    \label{fig:slice}
    \vspace{-3em}
\end{wrapfigure}

\Cref{tab:ram_accuracy} shows that RAM outperformed GGIK on all three metrics. While RAM was trained for a single epoch compared to 360 epochs for GGIK, its dataset was more than three orders of magnitude larger; despite this, RAM required less than a day of training, whereas GGIK required roughly a month. The majority of its inference time was due to transformations applied to its inputs and outputs, as these were neither batched nor accelerator-friendly. However, even the isolated model query took 83 microseconds, still about three orders of magnitude slower than RAM. RAM also achieved a balanced accuracy 15 percentage points higher, with a cross-seed confidence interval below 1 percentage point, indicating consistent training. The binary confusion matrix in \Cref{app:rq2} and the workspace slices in \Cref{fig:slice} show that the majority of the RAM-misclassified samples are false positives. This is consistent with the over-approximations in its training data and the implicit smoothing of the MLP architecture~\citep{SitzmannINR2020}. Crucially, a targeted evaluation of boundary poses in \Cref{app:boundary} roughly doubles the false positive rate, indicating a highly localised spatial dilation of the true reachable volume. Since RAM was specifically designed and trained for reachability, it solves a simpler problem: GGIK predicts link positions that are often physically infeasible and require solving for joint configurations that best match the predictions. This introduces additional deviations from the target pose and necessitates a high pose threshold of $\epsilon\approx0.2$, within which many unreachable poses have reachable neighbours.
\begin{table}[H]
    \begin{talltblr}[
        caption = {Comparison of RAM and GGIK. Inference is from pose and morphology input to reachability output.},
        label = {tab:ram_accuracy},
    ]{
        colspec = {l r r r},
        row{1} = {font=\bfseries}, 
    }
        \toprule
        Classifier & {Training  (h)} & {Inference (s)} & \SetCell{l}{Balanced (\%)\\ Accuracy }  \\
        \midrule
        RAM  & \textbf{21} & $\bm{147\cdot10^ {-9}}$ & {$\bm{84\,^{+0}_{-0}}$} \\
        GGIK & 756 & $22\cdot10^{-3}$ & 69 \\
        \bottomrule
    \end{talltblr}
\end{table}

\par
\begin{figure}[H]
    \begin{minipage}{0.67\linewidth}
        \centering
        \includegraphics[width=\linewidth]{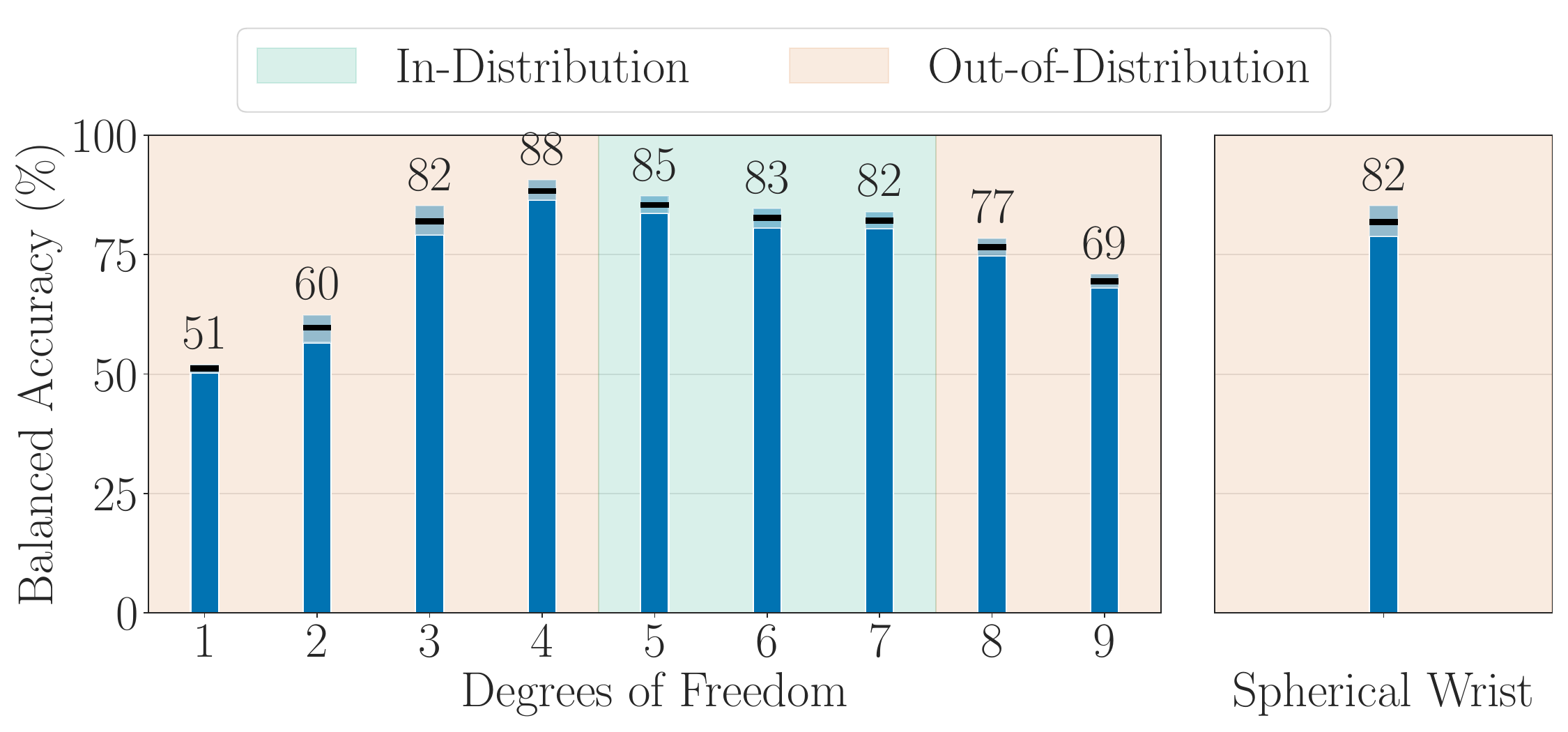}
        \caption{Balanced accuracy of RAM on morphologies in- and out-of-distribution.}
        \label{fig:ood}
    \end{minipage}
\end{figure}

\textbf{Out-of-distribution Generalisation.} To explore the limits of the generalisation of RAM, we evaluated it on morphologies with $\left\{1\text{-}4,8,9\right\}$ DoFs and morphologies with three consecutive co-located joints, i.e., spherical wrists. \Cref{fig:ood} shows that RAM roughly maintained in-distribution performance for three and four DoFs as well as for spherical wrists, with four DoFs even exceeding it by 4 percentage points. As shown by the binary confusion matrices in \Cref{app:generalisation}, accuracy collapsed for one- and two-DoF morphologies because the model predominantly predicted unreachable poses. Since most poses are unreachable for low-DoF morphologies, whose workspaces are lower-dimensional submanifolds of \SE, this is a graceful failure mode that limits false positives at the cost of recall. In contrast, accuracy degrades more gradually towards higher DoFs, suggesting that RAM increasingly fails to capture the reachable workspace, as evidenced by the falling true positive rate.
\par
\WFclear
\textbf{Cost of Generalisation.}
To isolate the cost of generalisation, we compared RAM with baselines specialised to each test morphology. These were reachability maps answering queries via direct cell lookup, i.e., the label-generation mechanism, and single-morphology RAMs (SMR), each trained on $10^6$ samples. \Cref{app:cost_of_cross} decomposes the 16-percentage-point gap between RAM and perfect balanced accuracy: 13 points stem from label quality, 1 point from approximating those labels with a neural network, and 2 points from generalising across morphologies.
\par
\textbf{Ablation and Exploration.}
We further examined how the granularity of discretisation affects the accuracy of RAM and probed the structure of its latent space. \Cref{app:granularity} confirms the practical ceiling from \Cref{ssec:rq1}, as accuracy degrades proportionally to label fidelity. \Cref{app:latent} finds that the latent space correlates most strongly with DoF, which the encoder receives directly as sequence length. The second strongest correlation is with the fraction of poses a morphology can reach, and morphologies close in latent space have similar workspaces, indicating that the encoder organises the latent space by reachability rather than by kinematic parameters alone.

\subsection{Utility of RAM (RQ3)}\label{ssec:rq3}
We performed experiments to evaluate how RAM affects the runtime and solution quality of morphology and trajectory optimisation compared to numerical inverse kinematics. Moreover, we isolated the effect of the accuracy and differentiability of RAM by optimising with CMA-ES~\citep{HansenCMA-ES2001, nomuraCMA-ES2024} and Adam~\citep{KingmaAdam2015}. Optimisations using RAM maximised the predicted reachability by optimising its inputs, a process pioneered in image classification~\citep{SimonyanDeepInside2014, DosovitskiyInvertingCNN2016, MahendranUnderstandCNN2015}, since its continuous output provides an informative gradient with respect to these inputs, in contrast to the binary indicator of \Cref{eq:exp_reachability} whose gradient is zero almost everywhere. Therefore, optimisations using numerical inverse kinematics had to minimise a smoothed objective, namely the sum of pose error and self-collisions, as detailed in \Cref{app:morphology_opt}. We obtained the gradients of this objective with respect to the optimised variables, i.e., the morphology parameters or the trajectory waypoints, by implicit differentiation through the numerical inverse kinematics solver~\citep{RaderOptimistix2024}.
\begin{table}[ht]
\centering
    \begin{talltblr}[
        caption = {Morphology optimisation with RAM compared to numerical inverse kinematics.},
label = {tab:rq3_morphology},
]{
    colspec = {l l r r r r},
row{1} = {font=\bfseries},
}
\toprule
Optimiser & Classifier & Runtime (s) & \SetCell{l}{Reached (\%)\\ Poses} & \SetCell{l}{Pose Error (\%)\\ Reduction} & \SetCell{l}{Self- (\%) \\ Collisions} \\
    \midrule
\SetCell[r=2]{l} Adam    & RAM & 0.1 & \num{79(2:1)} & \num{85(2:1)} & \num{3(1:0)} \\
                         & IK & 0.7 & \num{87(1:1)} & \num{93(1:1)} & $4\,^{+0}_{-1}$\\
    \midrule
\SetCell[r=2]{l} CMA-ES  & RAM & 0.3 & \num{86(2:1)} & \num{92(1:1)} & \num{5(1:1)} \\
                         & IK & 6.4 & \num{95(1:1)} & \num{97(1:0)} & $0\,^{+0}_{-0}$ \\
    \bottomrule
\end{talltblr}
\end{table}
\par
\textbf{Morphology Optimisation.} 
We maximised the reachability of random target poses by iteratively modifying the link lengths $\bm{a}$ and offsets $\bm{d}$ of an initial morphology, the only parameters amenable to gradient-based optimisation, since link twists are discrete. At each iteration, $\bm{a}$ and $\bm{d}$ were normalised, squashed, and renormalised to maintain a unit-size robot free of inevitable self-collisions, as detailed in \Cref{app:morphology_opt}. We optimised $1000$ random morphologies with six DoF, the minimum required to reach an arbitrary pose in \SE, over $100$ iterations towards $100$ task poses each.
\par
\begin{figure}
    \centering
    \subcaptionbox{RAM.\label{fig:design_opt_qualitative_ours}}{
        \centering\resizebox{0.25\linewidth}{!}{
        \begin{tikzpicture}
            \node
            at (0,0) {\includegraphics{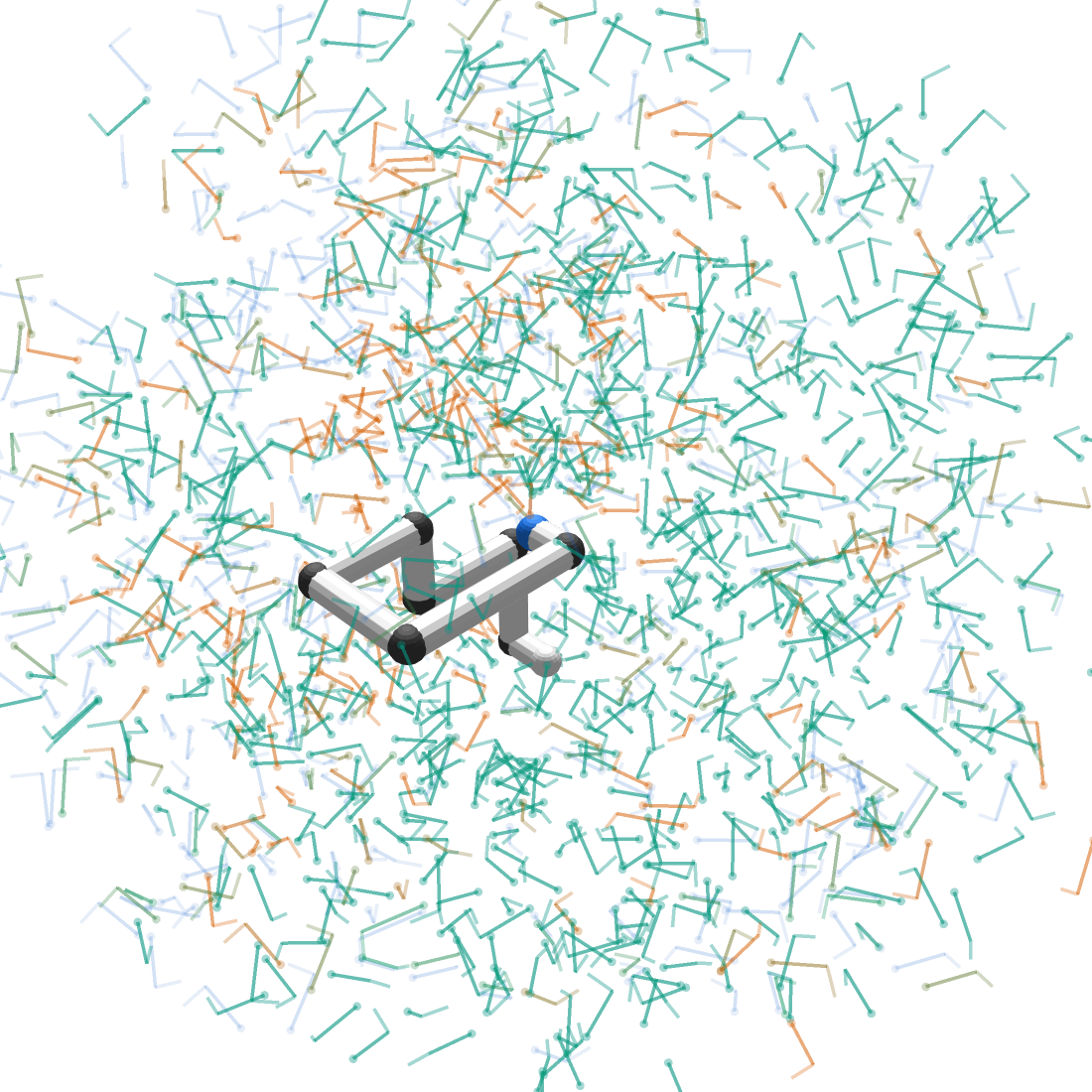}};
        \end{tikzpicture}
        }
    }\hfill
    \subcaptionbox{Initial.\label{fig:design_opt_qualitative_initial}}{
        \centering\resizebox{0.25\linewidth}{!}{
        \begin{tikzpicture}
            \node
            at (0,0) {\includegraphics{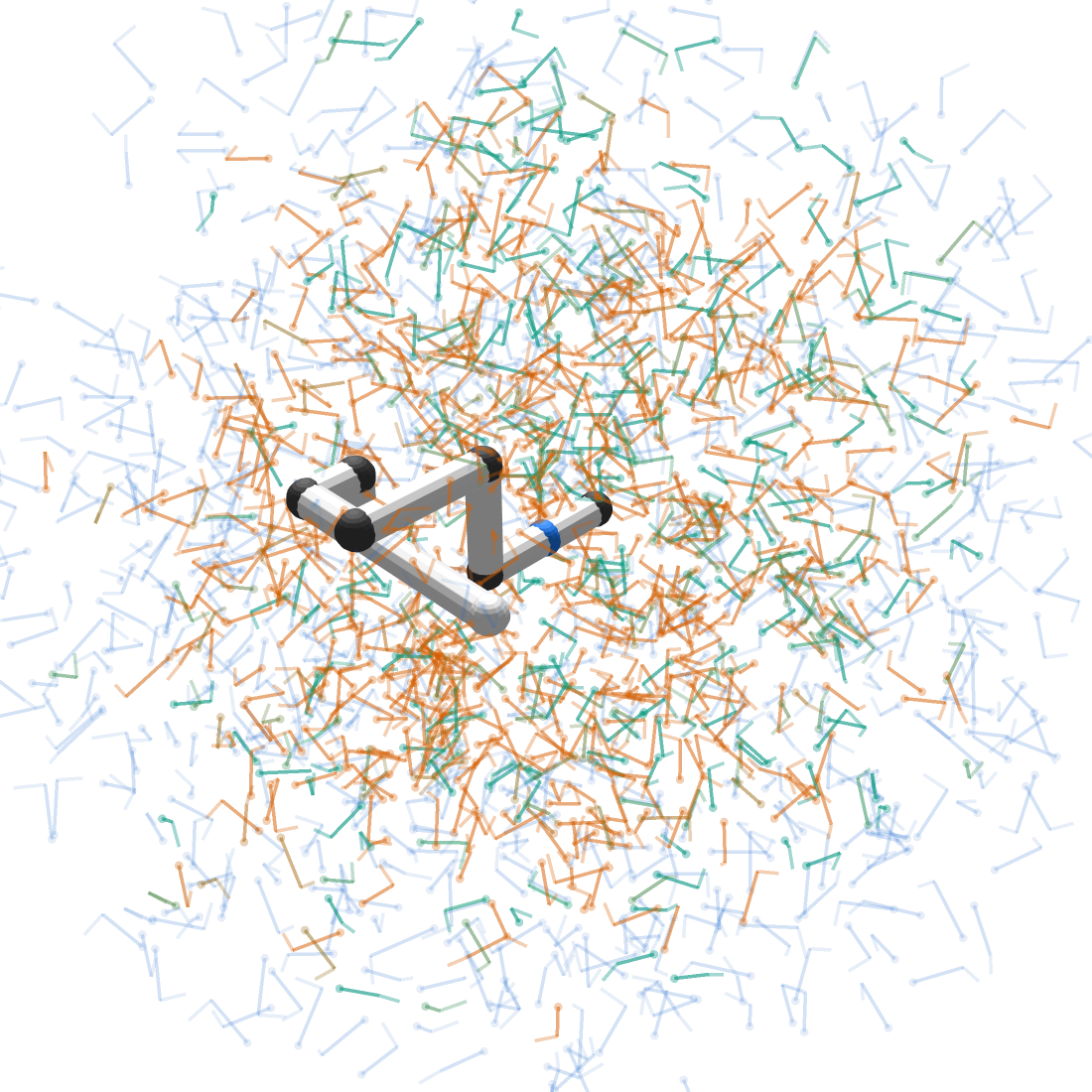}};
        \end{tikzpicture}
        }
    }\hfill
    \subcaptionbox{Inverse Kinematics.\label{fig:design_opt_qualitative_base}}{
        \centering\resizebox{0.25\linewidth}{!}{
        \begin{tikzpicture}
            \node
            at (0,0) {\includegraphics{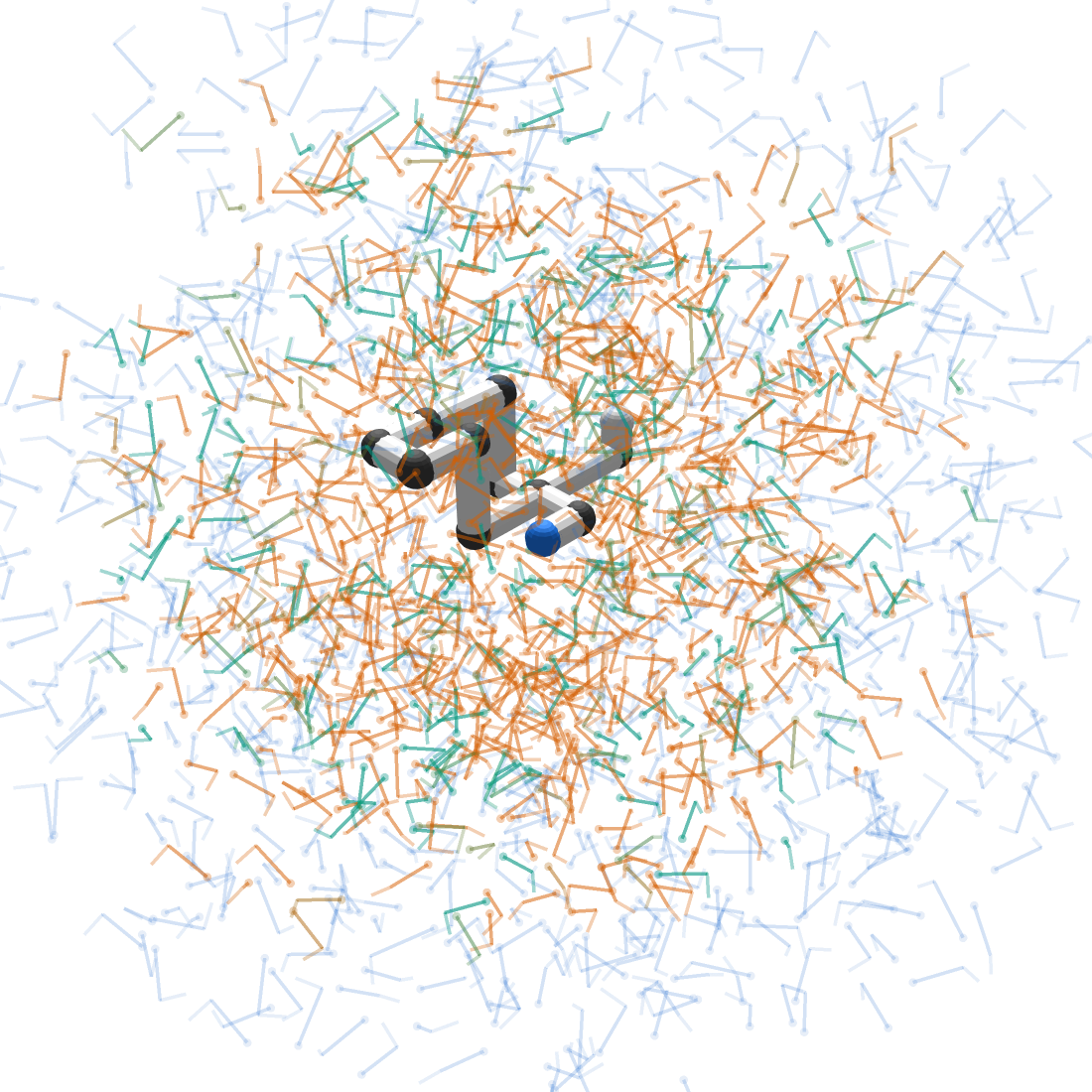}};
        \end{tikzpicture}
        }
    }
    \caption{Result of a morphology optimisation compared to the initial morphology. Task poses are displayed in opaque blue, whereas the actually reached poses fade from green (for close, reachable poses) to red for (poses far away or in self-collision).}
    \label{fig:design_opt_qualitative}
\end{figure}

\Cref{fig:design_opt_qualitative} shows an example in which both approaches produced substantial improvements over the initial morphology. Quantitatively, \Cref{tab:rq3_morphology} shows the same trade-off in both optimisers: RAM reduced runtime at a modest cost in reachability. The speed-up was $7\times$ for Adam and $21\times$ for CMA-ES, at the cost of 8 and 9 percentage points of reached poses, respectively. Since the reachability threshold $\epsilon$ is a design choice, we also measured the mean pose error, whose reduction trails the baseline by 8 and 5 percentage points. Self-collisions were the exception, where RAM outperformed the baseline under Adam.
\par
\textbf{Trajectory Optimisation} We also maximised the reachability along a nominal trajectory for a fixed morphology. Since adding a scaled gradient of the objective to a pose generally does not yield a valid pose (e.g., the rotation would no longer be orthogonal), we optimised for each waypoint a six-dimensional offset $\bm{v}\in\mathbb{R}^6$ instead, with three translational and three rotational components. We map the offset to a pose with the exponential map defined in \Cref{app:preliminaries}, $\bm{P}_{\text{opt}} = \exp(\bm{P}_{\text{nom}},\bm{v})$, so that any $\bm{v}$ yields a valid pose. We sampled the nominal trajectory from the geodesic between two random reachable poses and regularised the loss for deviations from the nominal trajectory $\anynorm{\bm{P}_{\text{opt}}-\bm{P}_{\text{nom}}}_{\SE}$. We optimised for 1000 different nominal trajectories, each described by 100 waypoints, over 100 iterations.
\par 
\Cref{app:trajectory_opt} shows that both approaches improved reachability by deviating from the nominal trajectory, with RAM again trading solution quality for runtime. The speed-up was $22\times$ for Adam and $60\times$ for CMA-ES, at the cost of 4 and 8 percentage points in the rate of reached poses and a three- to ten-fold larger deviation. The pose error reduction trails the baseline by 18 and 26 percentage points, which is consistent with the over-approximation of the training labels, drawing waypoints towards regions it predicts as reachable but which are not.

\section{Conclusion, Limitations, and Future Work}
RAM frames pose reachability as a binary occupancy classification problem, decoupling the existence of an inverse kinematics solution from the harder problem of finding them. RAM generalises across morphologies because the reachable workspace varies more smoothly with kinematic parameters than the solution set of inverse kinematics. Three limitations remain: RAM inherits the discretisation error of its training labels, establishing a ceiling on representation accuracy; it is restricted to conventional serial manipulators with 3-7 DoF and constant-radius capsule geometries; and it does not account for collisions with the environment. Advances in inverse kinematics solvers could provide higher-fidelity labels to overcome this discretisation ceiling. Beyond improving label quality, four directions stand out: conditioning the decoder on capsule radii to support varying link geometry, incorporating environmental obstacles as an additional model input, replacing the sequential encoder with a graph-based alternative to support arbitrary kinematic topologies, and exploiting the differentiable surrogate to jointly optimise morphology and trajectory in a single gradient-based loop, which the sequential nature of current design and planning pipelines precludes.

\clearpage
\acknowledgments{
This work was supported by the Deutsche Forschungsgemeinschaft (German Research Foundation) under grant number AL 1185/31-1, and the DASHH (HELMHOLTZ Graduate School for the Structure of Matter) under grant number HIDSS-0002. The authors gratefully acknowledge the computational and data resources provided by the Leibniz Supercomputing Centre (www.lrz.de). We would also like to thank Olivia Garland for the video voice-over and general advising.}

\bibliography{RAM}  

@article{AbdelMalekStratification2006,
  author = {{Abdel-Malek}, Karim and Yang, Jingzhou},
  journal = {The International Journal of Advanced Manufacturing Technology},
  number = {11},
  pages = {1211--1229},
  title = {Workspace Boundaries of Serial Manipulators Using Manifold Stratification},
  volume = {28},
  year = {2006},
}

@article{AmesIKFlow2022,
  author = {Ames, Barrett and Morgan, Jeremy and Konidaris, George},
  journal = {IEEE Robotics and Automation Letters},
  number = {3},
  pages = {7177--7184},
  title = {{IKFlow}: Generating Diverse Inverse Kinematics Solutions},
  volume = {7},
  year = {2022},
}

@inproceedings{BeesontracIK2015,
  author = {Beeson, Patrick and Ames, Barrett},
  booktitle = {Proc. of the IEEE-RAS Int. Conf. on Humanoid Robots (Humanoids)},
  pages = {928--935},
  title = {{TRAC-IK}: An Open-Source Library for Improved Solving of Generic Inverse Kinematics},
  year = {2015},
}

@inproceedings{BensadounIKNet2022,
  author = {Bensadoun, Raphael and Gur, Shir and Blau, Nitsan and Wolf, Lior},
  booktitle = {Proc. of the Int. Conf. on Machine Learning (ICML)},
  pages = {1787--1797},
  title = {Neural Inverse Kinematic},
  year = {2022},
}

@article{CavelliEllipsoid2025,
  author = {Cavelli, Rosario Francesco and Cen Cheng, Pangcheng David and Indri, Marina},
  journal = {Journal of Intelligent \& Robotic Systems},
  number = {3},
  title = {Modeling the {{Reachability Space}} of {{Robotic Manipulators}} through {{Ellipsoid Equations}}},
  volume = {111},
  year = {2025},
}

@article{ChirikjianConv1998,
  author = {Chirikjian, G.S. and {Ebert-Uphoff}, I.},
  journal = {IEEE Transactions on Robotics and Automation},
  number = {1},
  pages = {123--136},
  title = {Numerical Convolution on the {{Euclidean}} Group with Applications to Workspace Generation},
  volume = {14},
  year = {1998},
}

@article{ConkurRedundancy1997,
  author = {Conkur, E. Sahin and Buckingham, Rob},
  journal = {Robotica},
  number = {5},
  pages = {583--586},
  title = {Clarifying the Definition of Redundancy as Used in Robotics},
  volume = {15},
  year = {1997},
}

@book{CraigMDH2013,
  author = {Craig, John},
  publisher = {Pearson International},
  title = {Introduction to {{Robotics}}},
  year = {2013},
}

@article{DaiReconfiguration2024,
  author = {Dai, Ye and He, Shilong and Nie, XinLei and Rui, Xukun and Li, ShiKun and He, Sai},
  journal = {Journal of Intelligent \& Robotic Systems},
  number = {2},
  title = {Research on {{Reconfiguration Strategies}} for {{Self-reconfiguring Modular Robots}}: {{A Review}}},
  volume = {110},
  year = {2024},
}

@inproceedings{DerooEnergy2023,
  author = {Deroo, Boris and Pousett, Brendan and Aertbeli{\"e}n, Erwin and Decr{\'e}, Wilm and Bruyninckx, Herman},
  booktitle = {IEEE Int. Conf. on Automation Science and Engineering (CASE)},
  pages = {1--7},
  title = {The {{Need}} for {{Task-Specific Execution}} in {{Robot Manipulation}}: {{Skill Design}} for {{Energy-Efficient Control}}},
  year = {2023},
}

@phdthesis{DiankovIKFast2018,
  author = {Diankov, Rosen},
  school = {Carnegie Mellon University},
  title = {Automated Construction of Robotic Manipulation Programs},
  year = {2018},
}

@inproceedings{DosovitskiyInvertingCNN2016,
  author = {Dosovitskiy, Alexey and Brox, Thomas},
  booktitle = {Proc. of the IEEE Conf. on Computer Vision and Pattern Recognition (CVPR)},
  pages = {4829--4837},
  title = {Inverting Visual Representations with Convolutional Networks},
  year = {2016},
}

@article{FisherZ1915,
  author = {Fisher, R. A.},
  journal = {Biometrika},
  number = {4},
  pages = {507--521},
  title = {Frequency Distribution of the Values of the Correlation Coefficient in Samples from an Indefinitely Large Population},
  volume = {10},
  year = {1915},
}

@inproceedings{GiengerRandomforests2024,
  author = {Gienger, Andreas and Stein, Charlotte and Lauer, Anja P. R. and Sawodny, Oliver and Tar{\'i}n, Cristina},
  booktitle = {IEEE/SICE Int. Symp. on System Integration (SII)},
  pages = {1247--1252},
  title = {Data-{{Based Reachability Analysis}} and {{Optimized Robot Positioning}} for {{Co-Design}} of {{Construction Processes}}},
  year = {2024},
}

@article{HaDesignopt2018,
  author = {Ha, Sehoon and Coros, Stelian and Alspach, Alexander and Kim, Joohyung and Yamane, Katsu},
  journal = {The International Journal of Robotics Research},
  number = {13-14},
  pages = {1521--1536},
  title = {Computational Co-Optimization of Design Parameters and Motion Trajectories for Robotic Systems},
  volume = {37},
  year = {2018},
}

@inproceedings{HanSE2Conv2021,
  author = {Han, Yiheng and Pan, Jia and Xia, Mengfei and Zeng, Long and Liu, Yong-Jin},
  booktitle = {Proc. of the IEEE Int. Conf. on Robotics and Automation (ICRA)},
  pages = {1854--1860},
  title = {Efficient {{SE}}(3) {{Reachability Map Generation}} via {{Interplanar Integration}} of {{Intra-planar Convolutions}}},
  year = {2021},
}

@book{HartenbergDH1965,
  author = {Hartenberg, Richard Scheunemann and Denavit, Jacques},
  publisher = {McGraw-Hill},
  title = {Kinematic Synthesis of Linkages},
  year = {1964},
}

@article{HochreiterLSTM1997,
  author = {Hochreiter, Sepp and Schmidhuber, J{\"u}rgen},
  journal = {Neural Computation},
  number = {8},
  pages = {1735--1780},
  title = {Long {{Short-Term Memory}}},
  volume = {9},
  year = {1997},
}

@article{HoffmanBiLevel2025,
  author = {Hoffman, E. Mingo and Costanzi, D. and Fadini, G. and Miguel, N. and Prete, A. Del and Marchionni, L.},
  journal = {IEEE Robotics and Automation Letters},
  number = {3},
  pages = {2263--2270},
  title = {Addressing {{Reachability}} and {{Discrete Component Selection}} in {{Robotic Manipulator Design Through Kineto-Static Bi-Level Optimization}}},
  volume = {10},
  year = {2025},
}

@misc{IrshadNeuralFieldSurvey2024,
  author = {Irshad, Muhammad Zubair and Comi, Mauro and Lin, Yen-Chen and Heppert, Nick and Valada, Abhinav and Ambrus, Rares and Kira, Zsolt and Tremblay, Jonathan},
  howpublished = {arXiv:2410.20220},
  publisher = {arXiv},
  title = {Neural {{Fields}} in {{Robotics}}: {{A Survey}}},
  year = {2024},
}

@article{JaccardIoU1901,
  author = {Jaccard, Paul},
  journal = {Bull Soc Vaudoise Sci Nat},
  pages = {547--579},
  title = {{\'Etude} Comparative de La Distribution Florale Dans Une Portion Des {{Alpes}} et Des {{Jura}}},
  volume = {37},
  year = {1901},
}

@book{Johnson600-Cell2018,
  author = {Johnson, Norman W.},
  publisher = {Cambridge University Press},
  title = {Geometries and Transformations},
  year = {2018},
}

@inproceedings{KingmaAdam2015,
  author = {Kingma, Diederik P. and Ba, Jimmy},
  booktitle = {Proc. of the Int. Conf. on Learning Representations (ICLR)},
  title = {Adam: {{A}} Method for Stochastic Optimization},
  year = {2015},
}

@misc{KuelzCopilot2025,
  author = {K{\"u}lz, Jonathan and Ha, Sehoon and Althoff, Matthias},
  howpublished = {arXiv:2509.13077},
  publisher = {arXiv},
  title = {A {{Design Co-Pilot}} for {{Task-Tailored Manipulators}}},
  year = {2025},
}

@article{LeibrandtDiscretisation2023,
  author = {Leibrandt, Konrad and {da Cruz}, Lyndon and Bergeles, Christos},
  journal = {IEEE Transactions on Robotics},
  number = {4},
  pages = {2989--3007},
  title = {Designing {{Robots}} for {{Reachability}} and {{Dexterity}}: {{Continuum Surgical Robots}} as a {{Pretext Application}}},
  volume = {39},
  year = {2023},
}

@inproceedings{LiCDF2024,
  author = {Li, Yiming and Chi, Xuemin and Razmjoo, Amirreza and Calinon, Sylvain},
  booktitle = {Proc. of Robotics: Science and Systems (RSS)},
  title = {Configuration {{Space Distance Fields}} for {{Manipulation Planning}}},
  volume = {20},
  year = {2024},
}

@misc{LiSDFGeometry2024,
  author = {Li, Yiming and Zhang, Yan and Razmjoo, Amirreza and Calinon, Sylvain},
  howpublished = {arXiv:2307.00533},
  publisher = {arXiv},
  title = {Representing {{Robot Geometry}} as {{Distance Fields}}: {{Applications}} to {{Whole-body Manipulation}}},
  year = {2024},
}

@article{LimoyoGGIK2025,
  author = {Limoyo, Oliver and Mari{\'c}, Filip and Giamou, Matthew and Alexson, Petra and Petrovi{\'c}, Ivan and Kelly, Jonathan},
  journal = {IEEE Transactions on Robotics},
  pages = {1002--1018},
  title = {Generative {{Graphical Inverse Kinematics}}},
  volume = {41},
  year = {2025},
}

@book{LynchIK2017,
  author = {Lynch, Kevin M. and Park, Frank C.},
  publisher = {Cambridge University Press},
  title = {Modern Robotics},
  year = {2017},
}

@inproceedings{MahendranUnderstandCNN2015,
  author = {Mahendran, Aravindh and Vedaldi, Andrea},
  booktitle = {Proc. of the IEEE Conf. on Computer Vision and Pattern Recognition (CVPR)},
  pages = {5188--5196},
  title = {Understanding Deep Image Representations by Inverting Them},
  year = {2015},
}

@article{Mantel1967,
  author = {Mantel, Nathan},
  journal = {Cancer research},
  number = {2},
  pages = {209--220},
  title = {The Detection of Disease Clustering and a Generalized Regression Approach},
  volume = {27},
  year = {1967},
}

@inproceedings{MeschderOnet2019,
  author = {Mescheder, Lars and Oechsle, Michael and Niemeyer, Michael and Nowozin, Sebastian and Geiger, Andreas},
  booktitle = {Proc. of the IEEE Conf. on Computer Vision and Pattern Recognition (CVPR)},
  pages = {4460--4470},
  title = {Occupancy Networks: {{Learning {3D}}} Reconstruction in Function Space},
  year = {2019},
}

@misc{MildenhallNeRF2020,
  author = {Mildenhall, Ben and Srinivasan, Pratul P. and Tancik, Matthew and Barron, Jonathan T. and Ramamoorthi, Ravi and Ng, Ren},
  howpublished = {arXiv:2003.08934},
  publisher = {arXiv},
  title = {{NeRF}: Representing Scenes as Neural Radiance Fields for View Synthesis},
  year = {2020},
}

@article{MullerUniformPos1959,
  author = {Muller, Mervin E.},
  journal = {Communications of the ACM},
  number = {4},
  pages = {19--20},
  title = {A Note on a Method for Generating Points Uniformly on N-Dimensional Spheres},
  volume = {2},
  year = {1959},
}

@inproceedings{MurookaOnlyFK2025,
  author = {Murooka, Masaki and Kumagai, Iori and Morisawa, Mitsuharu and Kanehiro, Fumio},
  booktitle = {Proc. of the IEEE-RAS Int. Conf. on Humanoid Robots (Humanoids)},
  pages = {1--8},
  title = {Learning {{Differentiable Reachability Maps}} for {{Optimization-Based Humanoid Motion Generation}}},
  year = {2025},
}

@misc{OehsenLearnIK2020,
  author = {von Oehsen, Tim and Fabisch, Alexander and Kumar, Shivesh and Kirchner, Frank},
  howpublished = {arXiv:2003.00225},
  publisher = {arXiv},
  title = {Comparison of {{Distal Teacher Learning}} with {{Numerical}} and {{Analytical Methods}} to {{Solve Inverse Kinematics}} for {{Rigid-Body Mechanisms}}},
  year = {2020},
}

@inproceedings{Optuna,
  author = {Akiba, Takuya and Sano, Shotaro and Yanase, Toshihiko and Ohta, Takeru and Koyama, Masanori},
  booktitle = {Proceedings of the 25th ACM (SIGKDD) International Conference on Knowledge Discovery \& Data Mining},
  pages = {2623--2631},
  title = {Optuna: {{A Next-generation Hyperparameter Optimization Framework}}},
  year = {2019},
}

@article{OstermeierEAIK2025,
  author = {Ostermeier, Daniel and K{\"u}lz, Jonathan and Althoff, Matthias},
  journal = {IEEE Robotics and Automation Letters},
  number = {10},
  pages = {9964--9971},
  title = {Automatic {{Geometric Decomposition}} for {{Analytical Inverse Kinematics}}},
  volume = {10},
  year = {2025},
}

@inproceedings{ParkDeepSDF2019,
  author = {Park, Jeong Joon and Florence, Peter and Straub, Julian and Newcombe, Richard and Lovegrove, Steven},
  booktitle = {Proc. of the IEEE Conf. on Computer Vision and Pattern Recognition (CVPR)},
  pages = {165--174},
  title = {{DeepSDF}: Learning Continuous Signed Distance Functions for Shape Representation},
  year = {2019},
}

@article{ParkInvariantMetric1995,
  author = {Park, F. C.},
  journal = {Journal of Mechanical Design},
  number = {1},
  pages = {48--54},
  title = {Distance {{Metrics}} on the {{Rigid-Body Motions}} with {{Applications}} to {{Mechanism Design}}},
  volume = {117},
  year = {1995},
}

@book{PiperIK1969,
  author = {Pieper, Donald Lee},
  publisher = {Stanford University},
  title = {The Kinematics of Manipulators under Computer Control},
  year = {1969},
}

@article{QintToolUse2023,
  author = {Qin, Meiying and Brawer, Jake and Scassellati, Brian},
  journal = {Frontiers in Robotics and AI},
  title = {Robot Tool Use: {{A}} Survey},
  volume = {9},
  year = {2023},
}

@misc{RaderOptimistix2024,
  author = {Rader, Jason and Lyons, Terry and Kidger, Patrick},
  howpublished = {arXiv:2402.09983},
  publisher = {arXiv},
  title = {Optimistix: Modular Optimisation in {{JAX}} and {{Equinox}}},
  year = {2024},
}

@incollection{ReiterClosedworld1978,
  author = {Reiter, Raymond},
  booktitle = {Logic and {{Data Bases}}},
  editor = {Gallaire, Herv{\'e} and Minker, Jack},
  pages = {55--76},
  publisher = {Springer US},
  title = {On {{Closed World Data Bases}}},
  year = {1978},
}

@inproceedings{SchieleUnderactuated2022,
  author = {Schiele, Simon and Baumgartner, Sebastian and Laudahn, Simon and Lueth, Tim C.},
  booktitle = {Proc. of the IEEE/RSJ Int. Conf. on Intelligent Robots and Systems (IROS)},
  pages = {3047--3052},
  title = {Automated Design of Task Specific Additively Manufacturable Coupled Serial Chain Mechanisms for Tracing Predefined Planar Trajectories},
  year = {2022},
}

@inproceedings{SchuckScipy2025,
  author = {Schuck, Martin and {von Rohr}, Alexander and Schoellig, Angela P.},
  booktitle = {Proc. of the Int. Conf. on Neural Information Processing Systems (NeurIPS)},
  title = {Scipy.Spatial.Transform: {{Differentiable Framework-Agnostic {3D} Transformations}} in {{Python}}},
  year = {2025},
}

@inproceedings{vaswaniAttentionAllYou2017,
    author = {Vaswani, Ashish and Shazeer, Noam and Parmar, Niki and Uszkoreit, Jakob and Jones, Llion and Gomez, Aidan N. and Kaiser, {\L}ukasz and Polosukhin, Illia},
    booktitle = {Proc. of the Int. Conf. on Neural Information Processing Systems (NeurIPS)},
    title = {Attention Is All You Need},
    year = 2017,
}

@inproceedings{SeungsuFlow2021,
  author = {Kim, Seungsu and Perez, Julien},
  booktitle = {Proc. of the IEEE Int. Conf. on Robotics and Automation (ICRA)},
  pages = {4731--4737},
  title = {Learning {{Reachable Manifold}} and {{Inverse Mapping}} for a {{Redundant Robot}} Manipulator},
  year = {2021},
}

@article{ShirafujiKinematicSynthesis2025,
  author = {Shirafuji, Shouhei and Shimamura, Keiichiro},
  journal = {IEEE Robotics and Automation Letters},
  number = {3},
  pages = {2550--2557},
  title = {Kinematic {{Synthesis}} of a {{Serial Manipulator Using Gradient-Based Optimization}} on {{Lie Groups}}},
  volume = {10},
  year = {2025},
}

@incollection{ShoemakeUniformRandomRotations1992,
  author = {Shoemake, Ken},
  booktitle = {Graphics {{Gems III}}},
  pages = {124--132},
  publisher = {Academic Press Professional, Inc.},
  title = {Uniform Random Rotations},
  year = {1992},
}

@inproceedings{SimonyanDeepInside2014,
  author = {Simonyan, Karen and Vedaldi, Andrea and Zisserman, Andrew},
  booktitle = {Proc. of the Int. Conf. on Learning Representations (ICLR)},
  title = {Deep inside Convolutional Networks: {{Visualising}} Image Classification Models and Saliency Maps},
  year = {2014},
}

@inproceedings{SitzmannINR2020,
  author = {Sitzmann, Vincent and Martel, Julien N.P. and Bergman, Alexander W. and Lindell, David B. and Wetzstein, Gordon},
  booktitle = {Proc. of the Int. Conf. on Neural Information Processing Systems (NeurIPS)},
  pages = {7462--7473},
  title = {Implicit Neural Representations with Periodic Activation Functions},
  year = {2020},
}

@article{Spearman1904,
  author = {Spearman, C.},
  journal = {The American Journal of Psychology},
  number = {1},
  pages = {72--101},
  title = {The Proof and Measurement of Association between Two Things},
  volume = {15},
  year = {1904},
}

@article{HansenCMA-ES2001,
  author = {Hansen, N. and Ostermeier, A.},
  journal = {Evolutionary Computation},
  number = {2},
  pages = {159--195},
  title = {Completely Derandomized Self-Adaptation in Evolution Strategies},
  volume = {9},
  year = 2001,
}

@misc{nomuraCMA-ES2024,
  author={Masahiro Nomura and Masashi Shibata and Ryoki Hamano},
  howpublished = {arXiv:2402.01373},
  publisher = {arXiv},
  title = {cmaes: A Simple yet Practical Python Library for {CMA-ES}},
  year = {2024},
}

@inproceedings{SundaralingamCuRobo2023,
  author = {Sundaralingam, Balakumar and Hari, Siva Kumar Sastry and Fishman, Adam and Garrett, Caelan and Van Wyk, Karl and Blukis, Valts and Millane, Alexander and Oleynikova, Helen and Handa, Ankur and Ramos, Fabio and Ratliff, Nathan and Fox, Dieter},
  booktitle = {Proc. of the IEEE Int. Conf. on Robotics and Automation (ICRA)},
  pages = {8112--8119},
  title = {{CuRobo}: Parallelized Collision-Free Robot Motion Generation},
  year = {2023},
}

@inproceedings{VahrenkampVoxel2012,
  author = {Vahrenkamp, Nikolaus and Asfour, Tamim and Metta, Giorgio and Sandini, Giulio and Dillmann, R{\"u}diger},
  booktitle = {Proc. of the IEEE-RAS Int. Conf. on Humanoid Robots (Humanoids)},
  pages = {568--573},
  title = {Manipulability Analysis},
  year = {2012},
}

@inproceedings{Xinyu2024,
  author = {Chen, Xinyu and K{\"u}lz, Jonathan and Althoff, Matthias},
  booktitle = {Proc. of Robotics: Science and Systems (RSS)},
  title = {Generating Robot Capability Maps with Neural Fields},
  year = {2024},
}

@misc{YasutakeHJCDIK2025,
  author = {Yasutake, Cael and Kingston, Zachary and Plancher, Brian},
  howpublished = {arXiv:2510.07514},
  publisher = {arXiv},
  title = {{HJCD-IK}: {GPU}-Accelerated Inverse Kinematics through Batched Hybrid Jacobian Coordinate Descent},
  year = {2025},
}

@inproceedings{Yoshikawa1983,
  author = {Yoshikawa, Tsuneo},
  booktitle = {Proc. of the Int. Symp. on Robotics Research (ISRR)},
  pages = {735--747},
  title = {Analysis and Control of Robot Manipulators with Redundancy},
  year = {1983},
}

@inproceedings{ZachariasCapabilty2007,
  author = {Zacharias, Franziska and Borst, Christoph and Hirzinger, Gerd},
  booktitle = {Proc. of the IEEE/RSJ Int. Conf. on Intelligent Robots and Systems (IROS)},
  pages = {3229--3236},
  title = {Capturing Robot Workspace Structure: Representing Robot Capabilities},
  year = {2007},
}

@inproceedings{ZefranMetric1996,
  author = {{\v Z}efran, Milo{\v s} and Kumar, Vijay and Croke, Christopher},
  booktitle = {Proc. of the ASME Biennial Mechanisms Conf.},
  title = {Choice of {{Riemannian Metrics}} for {{Rigid Body Kinematics}}},
  year = {1996},
}

@inproceedings{ZhouContinuousrot2019,
  author = {Zhou, Yi and Barnes, Connelly and Lu, Jingwan and Yang, Jimei and Li, Hao},
  booktitle = {Proc. of the IEEE Conf. on Computer Vision and Pattern Recognition (CVPR)},
  pages = {5738--5746},
  title = {On the {{Continuity}} of {{Rotation Representations}} in {{Neural Networks}}},
  year = {2019},
}

\appendix

\setcounter{figure}{0} 
\renewcommand{\thefigure}{A\arabic{figure}}
\setcounter{table}{0} 
\renewcommand{\thetable}{A\arabic{table}}

\section{Preliminaries}\label{app:preliminaries}
Throughout the data generation process, we represent a pose with a homogeneous transformation matrix
\begin{equation}
    \bm{P} = \begin{bmatrix} \bm{R} & \bm{t} \\ \bm{0}^T & 1 \end{bmatrix} \in \SE \subset \mathbb{R}^{4\times4}
\end{equation}
with the rotation matrix $\bm{R} \in \SO \subset \mathbb{R}^{3\times3}$, translation vector $\bm{t} \in \mathbb{R}^3$, and zero vector $\bm{0}$. In this representation, forward kinematics is obtained by the cumulative matrix product of coordinate transformations for each link-joint pair and the end effector $f\left(\bm{M},\bm{\theta}\right) = \bm{P}_{\text{eef}} = \left(\prod_{i=0}^{n-1} \bm{T}_i \right) \bm{T}(\bm{m}_{\text{eef}}, 0)$, with the coordinate transformation of modified Denavit-Hartenberg parameters $\bm{T}_i$ defined as~\citep[Eq. 3.6]{CraigMDH2013}
\begin{equation}
    \label{eq:transformation_matrix}
         \bm{T}_i = \bm{T}\left(\bm{m}_i, \theta_i\right)  = \begin{bmatrix}
               \cos{\theta_i}                    & -\sin{\theta_i}                  & 0                   & a_{i-1}                \\
               \sin{\theta_i} \cos{\alpha_{i-1}} & \cos{\theta_i}\cos{\alpha_{i-1}} & -\sin{\alpha_{i-1}} & -d_i\sin{\alpha_{i-1}} \\
               \sin{\theta_i}\sin{\alpha_{i-1}}  & \cos{\theta_i}\sin{\alpha_{i-1}} & \cos{\alpha_{i-1}}  & d_i\cos{\alpha_{i-1}}  \\
               0                                 & 0                                & 0                   & 1
           \end{bmatrix} \, .
\end{equation}
Due to the use of finite-precision floating-point arithmetic, determining exact reachability is infeasible. Instead, we consider a pose reachable if some collision-free joint configuration places the end effector within a tolerance $\epsilon$ of it. This is formalised in \Cref{eq:exp_reachability} with $\mathcal{Q}=\mathcal{T}^n$, which cannot be evaluated exhaustively. For a finite $\mathcal{Q}\subset\mathcal{T}^n$ returned by a solver, a positive label certifies reachability, whereas a negative label may miss solutions. Since \SE lacks a canonical, bi-invariant metric~\citep{ParkInvariantMetric1995}, we weight translational and rotational errors~\citep[Theorem 4.4]{ZefranMetric1996}
\begin{equation}
    \anynorm{\bm{P}_1-\bm{P}_2}_{\SE} = \sqrt{c_1 \anynorm{\bm{t}_1-\bm{t}_2}_2^2 + c_2 \anynorm{\bm{R}_1^T\bm{R}_2}_{\SO}^2}\,,
\end{equation}
where $\anynorm{\cdot}_{\SO}$ denotes the geodesic distance. We set $c_1=\frac{1}{8}$ and $c_2=\frac{1}{2\pi^2}$ to bound both error components to $\frac{1}{2}$ for translations within the unit ball $\mathcal{B}^3$, ensuring both contribute equally to the metric. We set the tolerance to $\epsilon =10^{-4}$ to obtain reachability labels. 
\par
Since this metric treats translation and rotation independently, its geodesics combine a straight-line translation with a rotation about a fixed axis~\citep[Sec.~4.2]{ZefranMetric1996}. We generate them with the exponential and logarithmic maps. For a pose $\bm{P}$ and an offset $\bm{v}=(\bm{u},\bm{\omega})\in\mathbb{R}^6$ with translation vector $\bm{u}$ and rotation vector $\bm{\omega}$, the exponential map is
\begin{equation}
\exp(\bm{P},\bm{v}) = 
\begin{bmatrix} 
\bm{R}\,\mathrm{Rot}(\bm{\omega}) & \bm{t}+\bm{u} \\
\bm{0}^T & 1 
\end{bmatrix},
\end{equation}
where $\mathrm{Rot}(\bm{\omega})$ converts a rotation vector to its rotation matrix. The logarithmic map inverts it, 
\begin{equation}
    \log(\bm{P}_1,\bm{P}_2) = \begin{bmatrix}
        \bm{t}_2-\bm{t}_1 \\ \mathrm{Rot}^{-1}(\bm{R}_1^T\bm{R}_2)
    \end{bmatrix}\,,
\end{equation}
where $\mathrm{Rot}^{-1}$ returns the rotation vector with angle at most $\pi$. Scaling the offset, $\exp(\bm{P}_1, s\cdot\log(\bm{P}_1,\bm{P}_2))$ for $s\in[0,1]$, therefore traces the geodesic between $\bm{P}_1$ and $\bm{P}_2$, i.e., their shortest connection, which is unique if the relative rotation angle is below $\pi$.

\section{Data Synthesis}
This section provides further details on the data synthesis that was omitted from the main text.
\subsection{Morphology Sampling}\label{app:morph}
\begin{figure}[ht]
    \centering
    \definecolor{box}{RGB}{240,240,240}
    \definecolor{block}{RGB}{180,200,230}
    \definecolor{cardborder}{RGB}{200,200,200}
    \definecolor{SNSBlue}{HTML}{2496D5}

    \begin{tikzpicture}[
        node distance=0.3cm and 0.5cm,
        hero box/.style={
            rectangle, rounded corners=5pt, fill=box,
            inner sep=0pt, align=center
        },
        block style/.style={
            rectangle, draw=black!70, fill=block, rounded corners=2pt,
            inner sep=1pt, align=center, text=black!80,
        },
        decision/.style={
            diamond, draw=black!70, fill=white, aspect=1.5,
            inner sep=0pt, align=center, text=black!80,
            minimum width=2cm, minimum height=1cm
        },
        card label/.style={text=black!80, inner sep=1pt, align=center},
        arrow style/.style={->, >=latex, thin},
    ]

        \node[hero box, minimum width=\linewidth, minimum height=3cm] (container) {};

        \node[card label] (input) at ($(container.north west) + (0.75, -0.95)$) {$n$ \\$k$};
        \node[block style, right=of input, minimum width=2cm, minimum height=0.8cm] (s1) {Sample \\ Link Types};
        \node[decision, below=0.4cm of s1] (d1) {Reject};

        \node[block style, right=1.2cm of s1, minimum width=2cm, minimum height=0.8cm] (s2) {Sample \\ Link Twists};
        \node[card label] (link type input) at ($(s2.west) + (-0.3, 0.0)$) {$\bm{j}$};
        \node[decision, below=0.4cm of s2] (d2) {Reject};
        
        \node[block style, right=1.2cm of s2, minimum width=2cm, minimum height=0.8cm] (s3) {Sample \\ Link Lengths};
        \node[card label] (link type input2) at ($(s3.west) + (-0.3, 0.0)$) {$\bm{j}$};
        \node[decision, below=0.4cm of s3] (d3) {Reject};

        \node[decision, right=1.2cm of s3] (d4) {Reject};
        \node[card label] (morph input) at ($(d4.west) + (-0.3, 0.2)$) {$\bm{M}$};

        \node[card label] (output) at ($(d4.south) + (0.0, -0.75)$) {Morphology $\bm{M}$};
        
        \draw[arrow style] (input) -- (s1);
        
        \draw[arrow style] (s1) -- node[left] {$\bm{j}$} (d1);
        \draw[arrow style] (d1.west) -- ++(-0.4,0) |- node[pos=0.15, right] {Yes} ($(s1.west) +(0.0, -0.2)$);
        \draw[arrow style] (d1.east) -- ++(0.3,0) |- node[pos=0.1, left] {No} ($(s2.west) +(0.0, 0.2)$);
        
        \draw[arrow style] (s2) -- node[left] {$\bm{\alpha},\bm{j}$} (d2);
        \draw[arrow style] (d2.west) -- ++(-0.4,0) |- node[pos=0.15, right] {Yes} ($(s2.west) +(0.0, -0.2)$);
        \draw[arrow style] (d2.east) -- ++(0.3,0) |- node[pos=0.1, left] {No} ($(s3.west) +(0.0, 0.2)$);

        \draw[arrow style] (s3) -- node[left] {$\bm{a},\bm{d}$} (d3);
        \draw[arrow style] (d3.west) -- ++(-0.4,0) |- node[pos=0.15, right] {Yes} ($(s3.west) +(0.0, -0.2)$);
        \draw[arrow style] (d3.east) -- ++(0.5,0) |- node[pos=0.1, left] {No} (d4.west);
        
        \draw[arrow style] (d4.south) -- node[pos=0.1, left] {No} (output.north);
        \draw[arrow style] (d4.north) 
            -- ++(0, 0.2cm)              
            -- ++(-11.05cm, 0)             
            node[pos=0.0, right] {Yes} 
            |-                           
            ($(s1.west) + (0, 0.2)$);    

    \end{tikzpicture}
    \caption{Hierarchical rejection sampling of morphologies. Each stage samples new parameters and passes accumulated ones to the next. Variables on arrows denote dependencies of their destination.}
    \label{fig:sample_morph}
\end{figure}
The hierarchical rejection sampling scheme of \Cref{fig:sample_morph} receives the desired number of robots $k$ and the desired DoF $n$ as input. We first uniformly sample link types $\bm{j}\in\left\{0,1,2,3\right\}^{k\times (n+1)}$, which define the topology of a link, by the possession of a \textit{non-zero length} $a_{i-1}$ ($j=1$), a \textit{non-zero offset} $d_i$ ($j=2$), \textit{both} ($j=0$) length and offset, or \textit{neither}($j=3$) length nor offset; \Cref{fig:link_types} visualises examples. A link with neither offset nor length is essential for constructing co-located joints. Consecutive \textit{neither} link types are rejected as non-conventional. The second step uniformly samples link twists 
\begin{equation}
    \left(\bm{\alpha}\right)_{i,l} \in \begin{cases} 
         \left\{-\frac{\pi}{2}, 0 ,\frac{\pi}{2}\right\} & \text{if}\quad \left(\bm{j}\right)_{i,l}\neq 3\\
         \left\{-\frac{\pi}{2}, \frac{\pi}{2}\right\} & \text{else}
    \end{cases}    
\end{equation}
and rejects  more than three consecutive parallel or collinear axes as degenerate. The third step concerns link lengths $\bm{a}$ and offsets $\bm{d}$; it preserves distributional uniformity under the constraint $\sum_i\sqrt{a_i^2+d_i^2}=1$ by sampling as follows:
\begin{equation}
        \begin{bmatrix}
            \bm{a} \\ \bm{d}
        \end{bmatrix} =\frac{\bm{s}}{\anynorm{\bm{s}}_{\scalebox{0.75}{$1$}}}\begin{bmatrix}
            \sin\bm{\gamma} \\ \cos\bm{\gamma}
        \end{bmatrix},\quad\bm{s} \overset{\text{i.i.d.}}{\sim} \mathrm{Exponential}(1),\quad \bm{\gamma} \overset{\text{i.i.d.}}{\sim} \mathrm{Uniform}\left(0,2\pi\right),
\end{equation}
where $\mathrm{Exponential}(1)$ denotes the unit exponential distribution and $\mathrm{Uniform}\left(0,2\pi\right)$ the uniform distribution over $\left[0,2\pi\right]$. Link lengths and offsets are rejected if any element is non-zero and less than $2r$, as this would inevitably lead to so-called scissor self-collisions; these arise when adjacent capsules rotate into an anti-parallel configuration through a shared joint and overlap, as illustrated in \Cref{fig:joint_limits}. The final step rejects the complete morphology $\bm{M}$ if among $10^3$ randomly sampled joint configurations, all yield self-collisions or zero Yoshikawa manipulability indices~\citep{Yoshikawa1983}, indicating degeneracy. Some examples of resulting morphologies are displayed in \hyperref[fig:hero]{Figure 1-I}.

\subsection{Pose Sampling}\label{app:pose_sampling}
We sample poses uniformly from $\mathcal{B}^3(\bm{c}, r) \rtimes \SO$, with centre position $\bm{c}$ and radius $r$, by combining a uniformly sampled position $\bm{t} \in\mathcal{B}^3(\bm{c}, r)$ and a uniformly sampled orientation $\bm{R} \in \SO$. The position sampling is performed according to \citep{MullerUniformPos1959} as follows:
\begin{equation}
    \bm{t} = \bm{c}+\frac{\bm{d}}{\anynorm{\bm{d}}_2}\cdot r\sqrt[3]{u},\quad u\sim\mathrm{Uniform}\left(0,1\right),\quad\bm{d}\sim\mathrm{Normal}\left(\bm{0}_3, \bm{I}_3\right)\,,
\end{equation}
where $\bm{I}_3$ the $3\times3$ identity matrix. The orientation is sampled independently as a quaternion~\citep{ShoemakeUniformRandomRotations1992}
\begin{equation}
    \bm{q} = \frac{\bm{k}}{\anynorm{\bm{k}}_2} ,\quad \bm{k}\sim\mathrm{Normal}\left(\bm{0}_4, \bm{I}_4\right)
\end{equation}
and converted to a rotation matrix~\citep{SchuckScipy2025}. The resulting pose is composed with the end-effector transformation $\bm{T}(\bm{m}_{\text{eef}}, 0)$, which crucially does not require a joint configuration to obtain a geometrically plausible pose.

\subsection{Label Assignment}\label{app:label_assignment}
\begin{figure}[ht]
    \centering
    \begin{tikzpicture}[scale=0.25, transform shape]

        \definecolor{SNSBlue}{RGB}{1, 115, 178}
        \definecolor{SNSOrange}{RGB}{222, 143, 5}
        \definecolor{SNSRed}{RGB}{214, 39, 40}

        \def\r{2}
        \def\Lpre{16}
        \def\Lpost{8}
        \pgfmathsetmacro{\thetaarc}{asin(2*\r/\Lpost)}

        \coordinate (J)     at (0, 0);
        \coordinate (Jprev) at (-\Lpre, 0);

        \pgfmathsetmacro{\bndhi}{180 - \thetaarc}   
        \pgfmathsetmacro{\bndlo}{180 + \thetaarc}   

        \pgfmathsetmacro{\Bx}{\Lpost * cos(\bndhi)}   
        \pgfmathsetmacro{\By}{\Lpost * sin(\bndhi)}   
        \coordinate (Bnd) at (\Bx, \By);

        \pgfmathsetmacro{\operpx}{cos(\bndhi + 90) * \r}
        \pgfmathsetmacro{\operpy}{sin(\bndhi + 90) * \r}

        \fill[SNSRed!20]
            (J) -- (\bndhi:\Lpost)
            arc[start angle=\bndhi, end angle=\bndlo, radius=\Lpost]
            -- cycle;
        \draw[SNSRed, thick]
            (\bndhi:\Lpost) arc[start angle=\bndhi, end angle=\bndlo, radius=\Lpost];
        \node[SNSRed, scale=3.5, anchor=north]
            at ($(J)+(185:5.0)$) {\small forbidden};

        \draw[SNSOrange, thick]
            ($(J)+(\operpx, \operpy)$)   -- ($(Bnd)+(\operpx, \operpy)$);
        \draw[SNSOrange, thick]
            ($(J)+(-\operpx,-\operpy)$)  -- ($(Bnd)+(-\operpx,-\operpy)$);
        \draw[SNSOrange, thin, dashed]
            ($(J)$)  -- ($(Bnd)$);
        \draw[SNSOrange, thick] (Bnd) circle (\r);
        \node[SNSOrange, scale=3.5, anchor=south west]
            at ($(Bnd)+(1.5, 1.0)$) {$\bm{c}_\text{post}$};

        \draw[black!22, thin, dashed]
            (\bndlo:\Lpost)
            arc[start angle=\bndlo, end angle=\bndhi+360, radius=\Lpost];

        \draw[thick, dashed] (\Bx, \By) -- (\Bx, 0);
        \draw[thin] (\Bx-0.8,0)  -- (\Bx+0.8, 0);   
        \draw[thin] (\Bx-0.8,\By)-- (\Bx+0.8, \By);  
        \node[scale=3.5, anchor=west] at (\Bx-2.0, \By*0.3) {$2r$};

        \draw[SNSBlue, thick]
            ($(Jprev)+(0, \r)$)  -- ($(J)+(0, \r)$);
        \draw[SNSBlue, thick]
            ($(Jprev)+(0,-\r)$)  -- ($(J)+(0,-\r)$);
        \draw[SNSBlue, thin, dashed]
            ($(J)$)  -- ($(Jprev)$);
        \draw[SNSBlue, thick] (Jprev) circle (\r);
        \draw[SNSBlue, thick] (J)     circle (\r);
        \node[SNSBlue, scale=3.5, anchor=north] at (-9, {-\r-0.2})
            {$\bm{c}_\text{pre}$};

        \draw[black, thin, <->, >=stealth]
            ($(J)+(180:5.5)$)
            arc[start angle=180, end angle=\bndhi, radius=5.5];
        \node[scale=3.5, anchor=south east]
            at ($(J)+(173:3.0)$) {$\theta_\text{arc}$};

        \begin{scope}[shift={(J)}, scale=0.5]
            \fill[SNSBlue]
                (0,1) .. controls (0.05,0.05)  .. (1,0)
                      .. controls (0.05,-0.05) .. (0,-1)
                      .. controls (-0.05,-0.05).. (-1,0)
                      .. controls (-0.05,0.05) .. (0,1) -- cycle;
        \end{scope}

    \end{tikzpicture}
    \caption{Rotation-plane geometry for joint-limit derivation based on scissor-collision avoidance. The pre-capsule
    $\bm{v}_\text{pre}$ (blue) is fixed; the post-capsule $\bm{v}_\text{post}$
    (orange) is shown at the boundary of the forbidden arc (red, shaded), where
    its tip is exactly $2r$ from the axis of $\bm{v}_\text{pre}$. The forbidden
    arc spans $2\theta_\text{arc}$ centred on the anti-parallel direction; the
    joint is restricted to the complementary safe arc of span
    $2\pi - 2\theta_\text{arc}$.}
    \label{fig:joint_limits}
\end{figure}
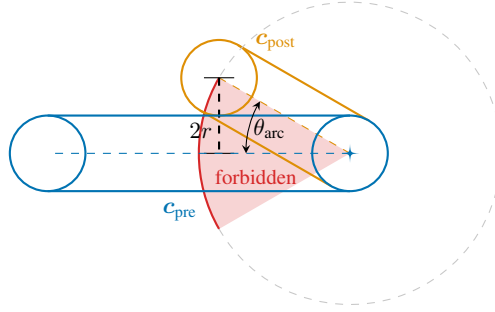

We sample joint configurations uniformly between joint limits $\theta_i \sim \mathrm{Uniform}\left[b_l, b_u\right]$, except for the end-effector angle $\theta_{\text{eef}}$, which is always zero as it is not followed by a subsequent joint. We derive the joint limits geometrically to avoid the scissor self-collisions depicted in \Cref{fig:joint_limits}, which remain the most common self-collisions, especially for capsules whose length is roughly $2r$.
For each joint $i$, let $\bm{v}_\text{pre}$ and $\bm{v}_\text{post}$ be the direction vectors of the nearest non-zero capsules proximal and distal to the joint, and $l_\text{min}$ the shorter of their lengths. A scissor collision requires both capsules to lie in the rotation plane of the joint: rotating about the joint axis does not change the tilt of a capsule relative to that plane, so a capsule outside the plane can never be folded back onto the other. The collision zone is centred on the configuration in which both capsules are anti-parallel. Collision begins once the tip of the shorter capsule comes within $2r$ of the axis of the other capsule, at a half-arc width of
\begin{equation}
    \theta_\text{arc} = \arcsin\left(\frac{2r}{l_\text{min}}\right)\,.
\end{equation}
For $l_\text{min} < 2r$, this holds at every joint angle, an inevitable self-collision, which motivates the constraint $a_{i-1}, d_i \in \{0\} \cup [2r, 1]$. Otherwise, the joint is restricted to the complementary arc of width $2\pi - 2\theta_\text{arc}$, resulting in $b_u = \phi_0 + \theta_\text{arc}$ and $b_l = b_u - (2\pi - 2\theta_\text{arc})$ with $\phi_0$ being the angle from $\bm{v}_\text{post}$ to the anti-parallel direction.

\section{Additional Results}
This section provides additional details on experiments in the main text as well as several new experiments and ablations. In subsequent tables, we abbreviate the True Positive Rate (TPR), False Negative Rate (FNR), False Positive Rate (FPR), and True Negative Rate (TNR).

\subsection{Label Fidelity (RQ1)}\label{app:rq1}
\Cref{tab:discretisation_fidelity_details} reports the binary confusion matrix for the various discretisation granularities.
\begin{table}[ht]
    \centering
    \caption{Binary confusion matrix of the discretisation granularities.}
    \label{tab:discretisation_fidelity_details}
    \begin{tblr}{
                colspec = {l r r r r},
                row{1} = {font=\bfseries},
            }
            \toprule
            Cell Spacing & TPR (\%) & FNR (\%) & FPR (\%) & TNR (\%)\\
            \midrule
            $\left[0.158, 0.163\right]$ & \num{100(0:0)} & \num{0(0:0)} & \num{59(4:4)} & \num{41(4:4)} \\
            $\left[0.080, 0.083\right]$ & \num{99(0:0)} & \num{1(0:0)} & \num{34(3:4)} & \num{66(4:3)} \\
            $\left[0.040, 0.042\right]$ & \num{97(1:0)} & \num{3(0:1)} & \num{23(3:4)} & \num{77(4:3)} \\
            $\left[0.021, 0.033\right]$ & \num{90(2:1)} & \num{10(1:2)} & \num{16(2:3)} & \num{84(3:2)} \\
            \bottomrule
    \end{tblr}
\end{table}

\subsection{Dataset Generation}\label{app:dataset}
The dataset comprises $3\cdot10^{10}$ samples from $3\cdot10^{4}$ morphologies, of which 29\% are reachable, totalling 204 GB generated in 360 GPU hours. About 38\% of sampled morphologies possess analytically solvable inverse kinematics. The vast majority of these have five DoF, as virtually all possess analytically solvable kinematics given our morphological constraints. To balance the reachability classes for these, we sample 25\% of their poses from the configuration space as $f\left(\bm{M}, \bm{\theta}\right)$ with $\bm{\theta}$ according to \Cref{app:label_assignment}.
\par
We empirically determined the number of samples per morphology to $10^{6}$, the threshold at which RAM, trained on a single morphology, generalised to unseen poses of that same morphology. The number of morphologies was then increased as much as possible within our computational budget, since generalisation to unknown morphologies increased monotonically with the amount of training data. Of the joint configurations sampled according to \Cref{app:label_assignment}, 85\% were collision-free, of which 23\% filled a previously empty cell, so 19\% of all forward-kinematics evaluations contributed to a new cell. The 19\% is an average over the entire sampling run; however, the fraction of configurations that fill a new cell is initially close to one and decreases as the workspace is covered, so most evaluations towards the end revisit cells that are already marked.

\subsection{Accuracy of RAM (RQ2)}\label{app:rq2}
\Cref{tab:rq2_binary_confusion} reports the binary confusion matrix for RAM and GGIK on the test set. Both methods misclassified far more unreachable poses as reachable than the reverse, but for different reasons: the error of RAM stems from the over-approximation inherent in its training labels and the implicit smoothing of its MLP decoder. The error of GGIK instead stems from its high pose-error threshold $\epsilon\approx0.2$, within which many genuinely unreachable poses have a reachable neighbour in its predicted solution set.
\begin{table}[ht]
    \centering
    \caption{Binary confusion matrix of RAM and GGIK.}
    \label{tab:rq2_binary_confusion}
    \begin{tblr}{
            colspec = {l r r r r},
            row{1} = {font=\bfseries}, 
        }
        \toprule
        Classifier & TPR (\%) & FNR (\%) & FPR (\%) & TNR (\%)\\
        \midrule
        RAM & \num{94(1:1)} & \num{6(1:1)} & \num{25(2:2)} & \num{75(2:2)}  \\
        GGIK & 90 & 10 & 61 & 39 \\
        \bottomrule
    \end{tblr}
\end{table}

\subsubsection{Workspace Slices}\label{app:slices}
We created the visualisation in \Cref{fig:slice} by fixing the orientation and one positional variable while varying the remaining two. To obtain slices that intersect the workspace, we started from a morphology $\bm{M}$ and computed its estimated workspace centre $\bm{t}_0$ and radius $r_\text{mov} = 1-\anynorm{\bm{t}_0}_2-\anynorm{\bm{t}_{\text{eef}}}_2$. We took the central position as the slice centre. The displayed orientation was chosen from the pose with the median Euclidean distance from the centre position. The displayed plane is the rotation plane of the first joint, which we determined by computing the transformation matrix $\bm{T}_0\left(\bm{m}_0, \bm{0}\right) = \begin{bmatrix}
    \bm{R}_0 & \bm{t}_0^T \\
    \bm{0} & 1
\end{bmatrix}$ and extracting the third column of its rotation matrix $\left(R_0\right)_{:, 3}$. We sampled each remaining positional variable with $10^3$ uniformly spaced values for a total of $10^6$ samples.

\subsubsection{Boundary Poses}\label{app:boundary}
Geometrically plausible poses were drawn from the whole reduced pose space, where most samples lie far from the reachability boundary. To probe the decision boundary directly, where the over-approximation of our training labels is most consequential, we also evaluated RAM and GGIK on poses near the boundary. We synthesised these by sampling $N$ equally spaced points on the geodesic between a reachable pose $\bm{P}_r$ and an unreachable pose $\bm{P}_u$, using the exponential and logarithmic maps defined in \Cref{app:preliminaries}:
\begin{equation}
\bm{P}_i = \exp\left(\bm{P}_r, \tfrac{i}{N}\bm{g}\right), \quad \bm{g} = \log\left(\bm{P}_r, \bm{P}_u\right), \quad i = 0,\dots,N.
\end{equation}
We evaluated RAM for 300 morphologies, $10^3$ pose pairs per morphology, and $N=100$. GGIK was again evaluated on only 1\% of the samples for each morphology.
\begin{table}[ht]
\centering
    \begin{talltblr}[
        caption = {Performance of RAM and GGIK on boundary poses.},
        label = {tab:ram_boundary},
    ]{
        colspec = {l r r r r r},
        row{1} = {font=\bfseries}, 
    }
        \toprule
        Classifier & Balanced Accuracy (\%)  & TPR (\%) & FNR (\%) & FPR (\%) & TNR (\%)  \\
        \midrule
        RAM & \textbf{\num{72(1:1)}}  & \num{94(1:1)} & \num{6(1:1)} & \num{49(2:2)} & \num{51(2:2)} \\
        GGIK & \num{62(1:1)}  & \num{64(3:4)} & \num{36(4:3)} & \num{39(2:2)} & \num{61(2:2)} \\
        \bottomrule
    \end{talltblr}
\end{table}
\par
\Cref{tab:ram_boundary} shows that RAM also outperformed GGIK on boundary poses by 10 percentage points, though both methods lost accuracy relative to \Cref{tab:ram_accuracy}. The two methods failed in opposite directions: GGIK was conservative, with a higher true negative rate but missing over a third of reachable poses, whereas RAM retained a high true positive rate at the cost of a false positive rate that increased from 25\% on random poses to 49\% on boundary poses. The errors of RAM are thus concentrated at the boundary and consist of false positives inherited from the over-approximated training labels, consistent with a dilated version of the workspace.

\subsubsection{Out-of-distribution Generalisation}\label{app:generalisation}
\Cref{tab:generalisation} reports the balanced accuracy and the binary confusion matrix underlying \Cref{fig:ood}. All results in this experiment were averaged over 100 morphologies, each with $10^5$ pose samples. From one to six DoFs, both the true and false positive rates increased monotonically: RAM became steadily less conservative as the DoF increased, with no abrupt change at the boundary of the training distribution. Three DoF morphologies illustrate this, reaching a balanced accuracy comparable to the in-distribution range but with a significantly lower true positive rate and false positive rate than any in-distribution DoF.
\begin{table}[ht]
\centering
    \begin{talltblr}[
        caption = {Generalisation capabilities of RAM.},
        label = {tab:generalisation},
    ]{
        colspec = {l r r r r r},
        row{1} = {font=\bfseries}, 
    }
        \toprule
        Data & Balanced Accuracy (\%) & TPR (\%) & FNR (\%) & FPR (\%) & TNR (\%)\\
        \midrule

1 DoF & \num{51(1:1)} & \num{3(2:3)} & \num{93(5:4)} & \num{0(1:0)} & \num{100(0:1)}\\
2 DoF & \num{60(2:3)} & \num{22(6:7)} & \num{78(7:6)} & \num{3(0:2)} & \num{97(2:0)}\\
3 DoF & \num{82(3:3)} & \num{72(6:6)} & \num{28(6:6)} & \num{8(1:2)} & \num{92(2:1)}\\
4 DoF & \num{88(3:2)} & \num{89(4:4)} & \num{11(4:4)} & \num{12(2:1)} & \num{88(1:2)}\\
\midrule
5 DoF & \num{85(2:1)} & \num{92(3:3)} & \num{8(3:3)} & \num{21(2:2)} & \num{79(2:2)}\\
6 DoF & \num{83(2:2)} & \num{95(1:1)} & \num{5(1:1)} & \num{29(4:3)} & \num{71(3:4)}\\
7 DoF & \num{82(2:2)} & \num{90(1:2)} & \num{10(2:1)} & \num{25(3:3)} & \num{75(3:3)}\\
\midrule
8 DoF & \num{77(1:2)} & \num{76(2:2)} & \num{24(2:2)} & \num{23(2:3)} & \num{77(3:2)}\\
9 DoF & \num{69(2:1)} & \num{59(2:2)} & \num{41(2:2)} & \num{20(2:2)} & \num{80(2:2)}\\
\midrule
Spherical Wrist & \num{82(3:3)} & \num{84(6:5)} & \num{9(2:4)} & \num{21(2:3)} & \num{79(3:2)}\\

        \bottomrule
    \end{talltblr}
\end{table}

\subsubsection{Cost of Generalisation}\label{app:cost_of_cross}
For each of the $300$ test morphologies, we built a reachability map with the workspace approximation used to generate the training data, and trained a separate RAM instance on $10^6$ samples generated solely from that morphology. Each baseline was evaluated only on the test samples of its own morphology. \Cref{tab:ram_cost_of_cross} shows that the false positive rate is similar across all three classifiers, indicating that the false positives of RAM are almost entirely inherited from the over-approximation of its training labels rather than introduced by learning or generalisation. The cost of generalisation instead appears mostly in the true positive rate, which drops from 97\% for the reachability map to 93\% for RAM.
\begin{table}[ht]
\centering
    \begin{talltblr}[
        caption = {Comparison of RAM to single-morphology baselines.},
        label = {tab:ram_cost_of_cross},
    ]{
        colspec = {l r r r r r},
        row{1} = {font=\bfseries}, 
    }
        \toprule
        Classifier & {Balanced (\%)\\Accuracy}  & TPR (\%) & FNR (\%) & FPR (\%) & TNR (\%)  \\
        \midrule
        RAM & \num{84(1:1)}  & \num{93(1:1)} & \num{7(1:1)} & \num{25(2:1)} & \num{75(1:2)} \\
        SMR & \num{86(1:1)}  & $96\,^{+0}_{-1}$ & $4\,^{+1}_{-0}$ & \num{23(2:1)} & \num{77(1:2)} \\
        Reachability Map & \num{87(2:2)}  & \num{97(1:0)} & $3\,^{+0}_{-1}$ & \num{23(3:4)} & \num{77(4:3)} \\
        \bottomrule
    \end{talltblr}
\end{table}

\subsection{Utility of RAM (RQ3)}
This section states the optimisation objectives for both tasks and reports their convergence behaviour over iterations.

\subsubsection{Morphology Optimisation}\label{app:morphology_opt}
\begin{figure}[H]
    \centering
    \includegraphics[width=\linewidth]{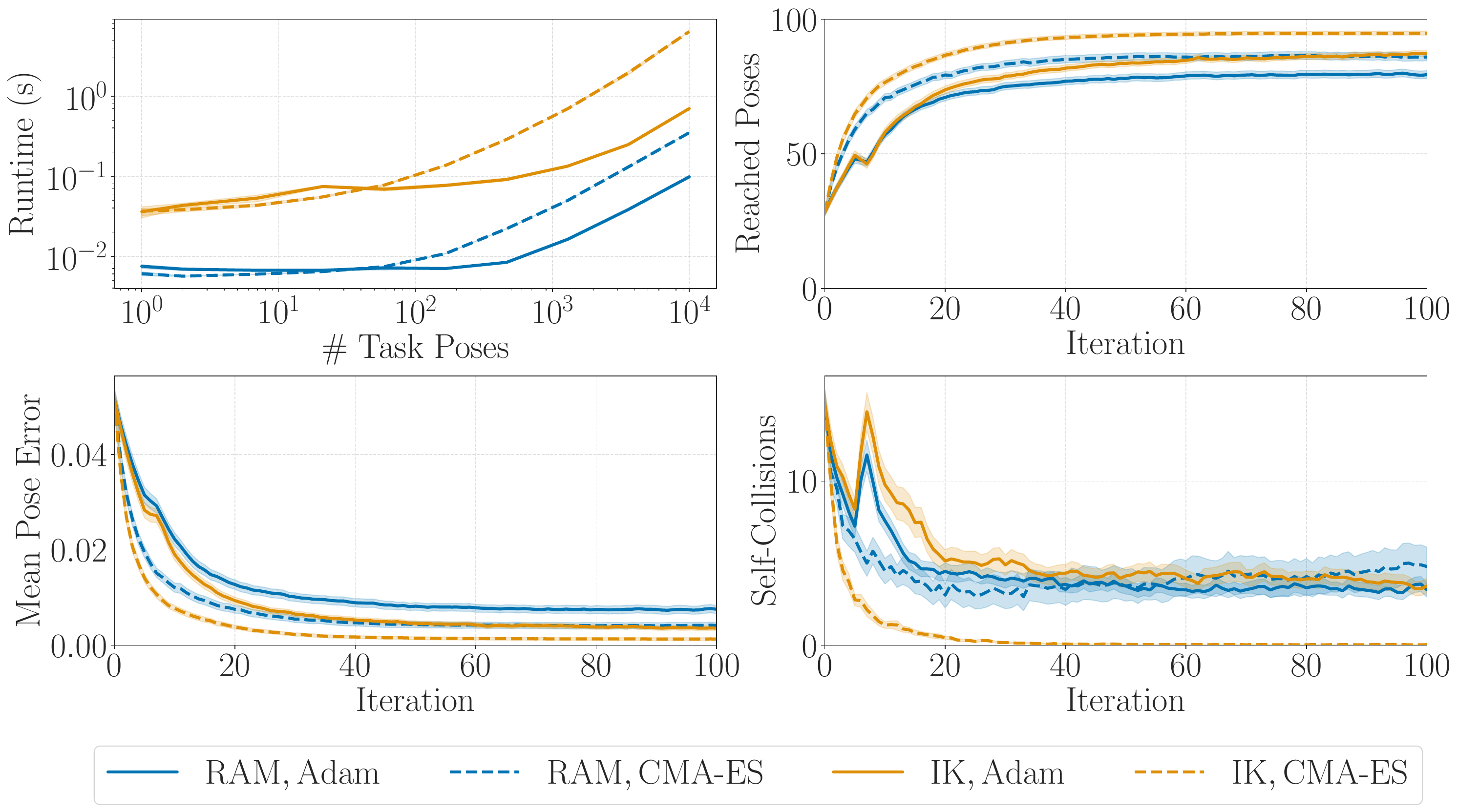}
    \caption{Comparison of morphology-optimisation approaches with regards to quality and runtime.}
    \label{fig:design_opt}
\end{figure}

We optimised the morphology $\bm{M}$ by maximising predicted reachability over a set of $N$ random target poses $\{\bm{p}_0, \dots, \bm{p}_N\}$:
\begin{equation}\label{eq:morphology_objective}
\max_{\bm{M}}\frac{1}{N}\sum_i \text{RAM}\left(\bm{M},\bm{p}_i\right).
\end{equation}
The baseline instead performs the following minimisation: 
\begin{equation}\label{eq:morphology_objective2}
\min_{\bm{M}}\frac{1}{N}\sum_i
\max\left(0,\anynorm{\bm{P}_i - f\left(\bm{M}, f^{-1}\left(\bm{M},\bm{P}_i\right)\right)}_{\SE}-\epsilon\right)
+ \lambda c\left(\bm{M}, f^{-1}\left(\bm{M},\bm{P}_i\right)\right),
\end{equation}
with gradients from the implicit function theorem applied to the solver, the \SE tolerance $\epsilon=10^{-4}$ from \Cref{app:preliminaries}, and a penalty weight $\lambda=10^4$ chosen to dominate the pose-error term whenever a self-collision occurs. At each iteration, we normalised $\bm{a}$ and $\bm{d}$ to unit size, squashed any element below $2r$ to zero to prevent inevitable self-collisions, and renormalised. Since squashing only shrinks parameters, the second normalisation can only increase them and therefore cannot require further squashing. Squashing is not differentiable and was disregarded in the backward pass, so the gradients are approximate near the threshold. 
\Cref{fig:design_opt} shows how the quantities of \Cref{tab:rq3_morphology} evolve over the optimisation, as well as the runtime scaling.

\subsubsection{Trajectory Optimisation}\label{app:trajectory_opt}
\begin{figure}[H]
    \centering
    \includegraphics[width=\linewidth]{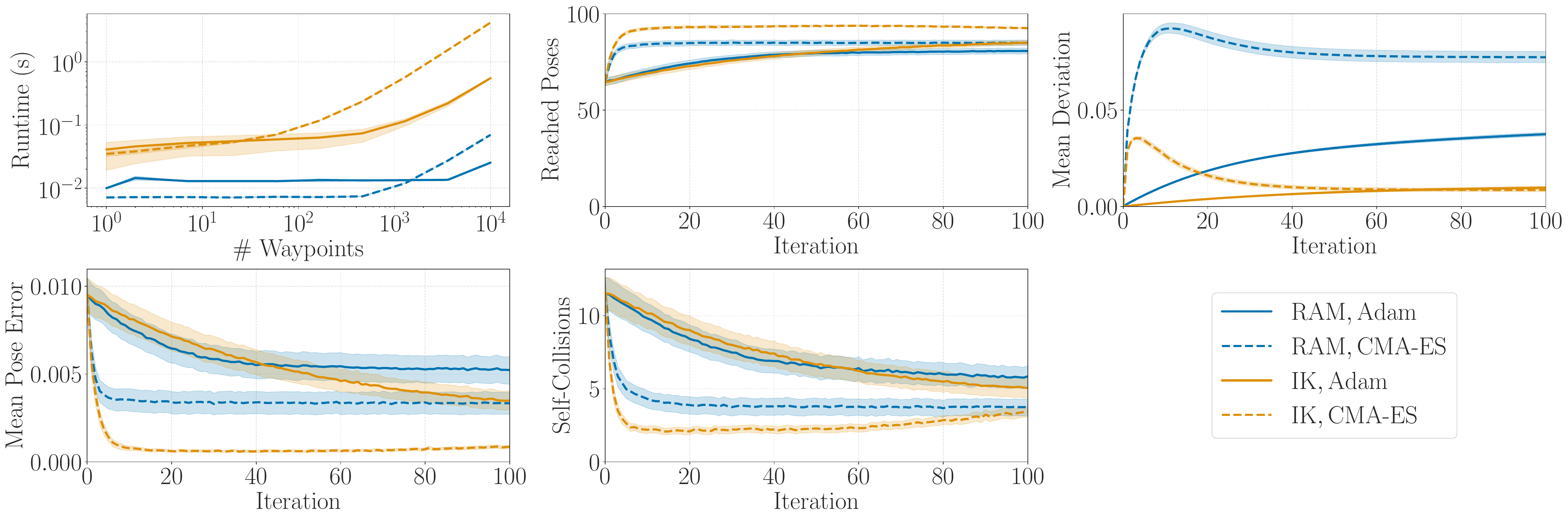}
    \caption{Comparison of trajectory-optimisation approaches with regards to quality and runtime.}
    \label{fig:trajectory_optimisation}
\end{figure}

We optimised the trajectory over the tangent vectors $\bm{V}$ for a fixed morphology $\bm{M}$, maximising predicted reachability while penalising deviations from the nominal trajectory as
\begin{equation}
\max_{\bm{V}}\frac{1}{N}\sum_i \text{RAM}\left(\bm{M},\bm{p}_i\right) - \mu\anynorm{\bm{P}_{i}-\bm{P}_{\text{nom},i}}_{\SE}\,,
\end{equation}
where  $\bm{P}_i = \exp\!\left(\bm{P}_{\text{nom},i},\bm{v}_i\right)$, $\bm{p}_i$ is the vectorised representation of $\bm{P}_i$, $\bm{V}$ collects the tangent vectors $\{\bm{v}_i\}$ into a matrix, and $\bm{P}_{\text{nom},i}$ is the $i$-th waypoint on the nominal trajectory. The baseline instead solves the following minimisation:
\begin{equation}
\begin{aligned}
\min_{\bm{V}}\frac{1}{N}\sum_i
&\max\left(0,\anynorm{\bm{P}_i - f\left(\bm{M}, f^{-1}\left(\bm{M},\bm{P}_i\right)\right)}_{\SE}-\epsilon\right)\\
&+ \lambda c\left(\bm{M}, f^{-1}\left(\bm{M},\bm{P}_i\right)\right)
+ \mu\anynorm{\bm{P}_{i}-\bm{P}_{\text{nom},i}}_{\SE}\,.
\end{aligned}
\end{equation}
We reused $\epsilon$ and $\lambda$ from morphology optimisation and empirically chose the deviation weight $\mu=\frac{1}{20}$. \Cref{tab:rq3_trajectory} reports the results, and \Cref{fig:trajectory_optimisation} shows how they evolve over the optimisation.

\begin{table}[ht]
\centering
    \begin{talltblr}[
        caption = {Trajectory optimisation with RAM compared to numerical inverse kinematics.},
label = {tab:rq3_trajectory},
]{
    colspec = {l r r r r },
row{1-2} = {font=\bfseries},
}
\toprule
Optimiser & \SetCell[c=2]{c} Adam & & \SetCell[c=2]{c} CMA-ES & \\
\cmidrule[lr]{2-3} \cmidrule[lr]{4-5}
Classifier & RAM & IK & RAM & IK \\
    \midrule
Runtime (s)                & 0.025 & 0.55 & 0.07 & 4.2 \\
Reached Poses (\%)         & \num{81(1:2)} & \num{85(1:1)} & \num{85(1:1)} & \num{93(0:1)} \\
Relative Mean Deviation    & \num{1.0(0.0:0.0)}$\times$ & \num{0.3(0.0:0.1)}$\times$ & \num{2.1(0.0:0.1)}$\times$ & \num{0.2(0.1:0.0)}$\times$ \\
Pose Error Reduction (\%)  & \num{45(8:8)} & \num{63(6:5)} & \num{65(7:7)} & \num{91(1:1)} \\
Self-Collisions (\%)       & \num{6(1:1)} & \num{5(1:1)} & \num{4(0:1)} & \num{3(1:0)} \\
    \bottomrule
\end{talltblr}
\end{table}

\subsection{Ablations}
This section adds further details to the ablation studies.

\subsubsection{Impact of the Discretisation Granularity on RAM}\label{app:granularity}
To investigate the effect of training label accuracy on RAM, we generated two additional datasets of identical size at the lower discretisation granularities of \Cref{tab:discretisation_fidelity} and trained RAM on each over ten seeds, evaluating against the same test set as in \Cref{ssec:rq2}. \Cref{tab:ram_discretisation} shows that the balanced accuracy of RAM closely tracks that of its training labels, trailing it by 2 to 4 percentage points throughout. Coarser labels, therefore, lead to lower accuracy, confirming that label fidelity limits representation accuracy.
\begin{table}[ht]
\centering
    \begin{talltblr}[
        caption = {RAMs trained on different training label fidelities.},
        label = {tab:ram_discretisation},
    ]{
        colspec = {l r r},
        row{1} = {font=\bfseries}, 
    }
        \toprule
        Cell Spacing & \SetCell{l}{Training Labels \\ Balanced Accuracy (\%)} & \SetCell{l}{RAM \\ Balanced Accuracy (\%)}\\
        \midrule
        $\left[0.158, 0.163\right]$ & \num{71(2:3)} & \num{69(1:3)} \\
        $\left[0.080, 0.083\right]$ & \num{83(2:2)} & \num{79(3:1)} \\
        $\left[0.040, 0.042\right]$ & \num{87(2:2)} & $84\,^{+0}_{-0}$ \\
        \bottomrule
    \end{talltblr}
\end{table}

\subsubsection{Encoder Architecture}\label{app:transformer}
To validate our choice of encoder, we replaced the LSTM with a transformer~\citep{vaswaniAttentionAllYou2017} and retrained under otherwise equivalent hyperparameters, without separate tuning. \Cref{tab:ram_transformer} shows that the LSTM outperformed the transformer on all three metrics, with a balanced accuracy 9 percentage points higher at roughly a quarter of the inference time. We attribute this to the short sequences considered (at most eight tokens), for which self-attention offers little advantage over sequential processing while incurring a higher per-query cost.
\begin{table}[ht]
\centering
    \begin{talltblr}[
        caption = {Comparison of encoder types for RAM.},
        label = {tab:ram_transformer},
    ]{
        colspec = {l r r r},
        row{1} = {font=\bfseries}, 
    }
        \toprule
        Classifier & {Training  (h)} & {Inference (s)} & \SetCell{l}{Balanced (\%)\\ Accuracy }  \\
        \midrule
        LSTM  & \textbf{21} & $\bm{147\cdot10^ {-9}}$ & $\bm{84\,^{+0}_{-0}}$ \\
        Transformer  & 25 & $553\cdot10^{-9}$ & \num{75(1:1)} \\
        \bottomrule
    \end{talltblr}
\end{table}

\subsection{Latent Space Exploration}\label{app:latent}
We investigated whether the latent space captures interpretable morphological properties and whether distances between latent representations reflect structural and functional similarities between morphologies. First, we sampled $3\cdot10^3$ morphologies, extracted their final LSTM hidden-state vectors, and projected these representations onto their principal components. For each morphological attribute, we computed the Spearman correlation coefficient~\citep{Spearman1904} and the corresponding $p$-value for each principal component, retaining the component with the largest absolute correlation. 95\% confidence intervals were computed using the Fisher-$z$ transform~\citep{FisherZ1915}. We considered attributes describing kinematic complexity, geometry, workspace properties, and joint composition, with \Cref{tab:exploration} reporting coefficients, confidence intervals, and $p$-values for all of them. The strongest correlations were with DoF and reachability, at 87\% and 65\% respectively, indicating that the latent representation captures not only kinematic structure but already an aggregate measure of workspace volume.
\par
\begin{figure}
 \centering
\subcaptionbox{Correlation of DoF}[0.5\linewidth]{\centering\includegraphics[height=2.5cm]{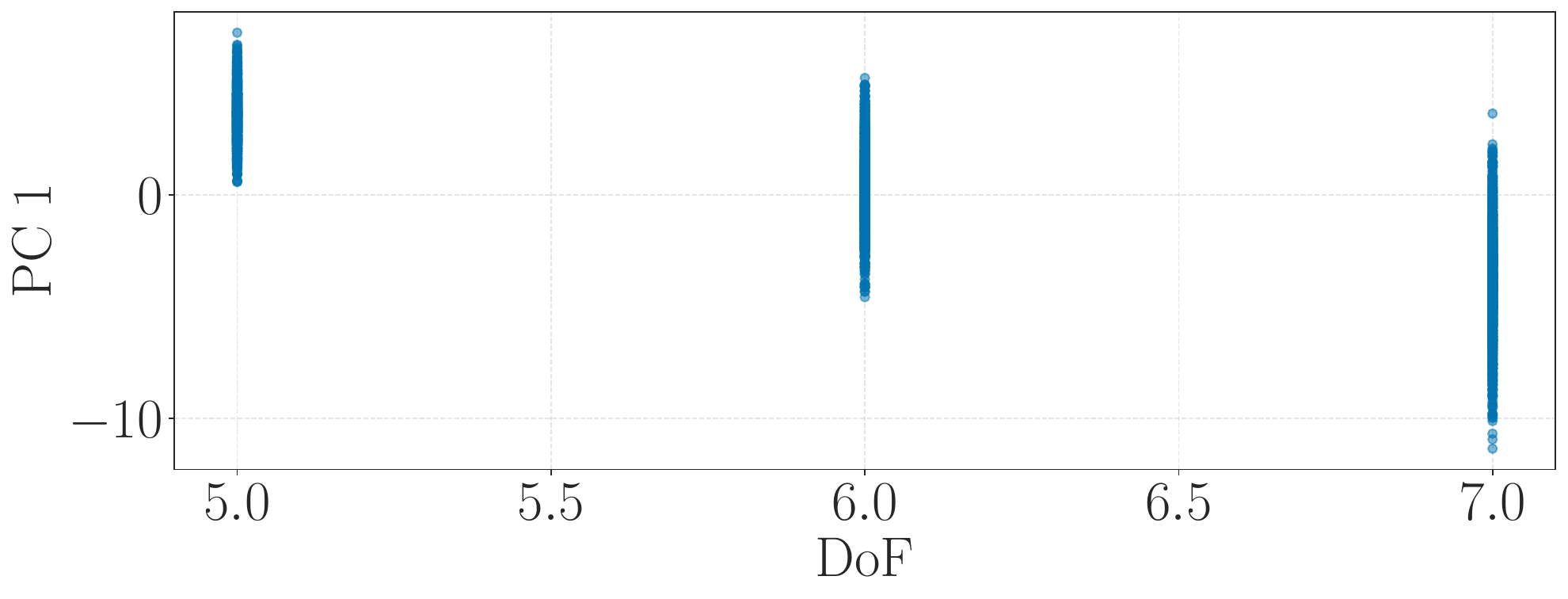}}\hfill
 \subcaptionbox{Correlation of reachability $\frac{\sum^{1000}_il_i}{1000}$}[0.5\linewidth]{\centering\includegraphics[height=2.5cm]{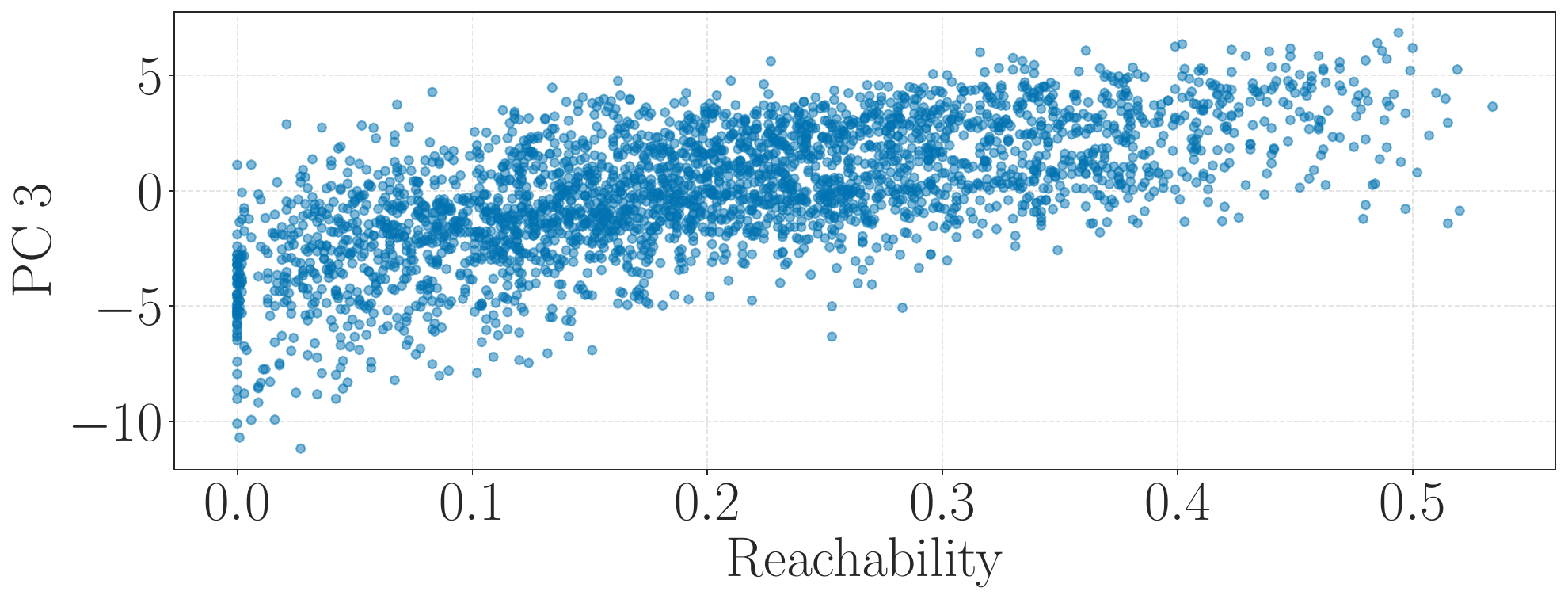}}\hfill
 \subcaptionbox{$d_\text{MDH}$ vs. $d_\text{latent}$}[0.5\linewidth]{\centering\includegraphics[scale=0.2]{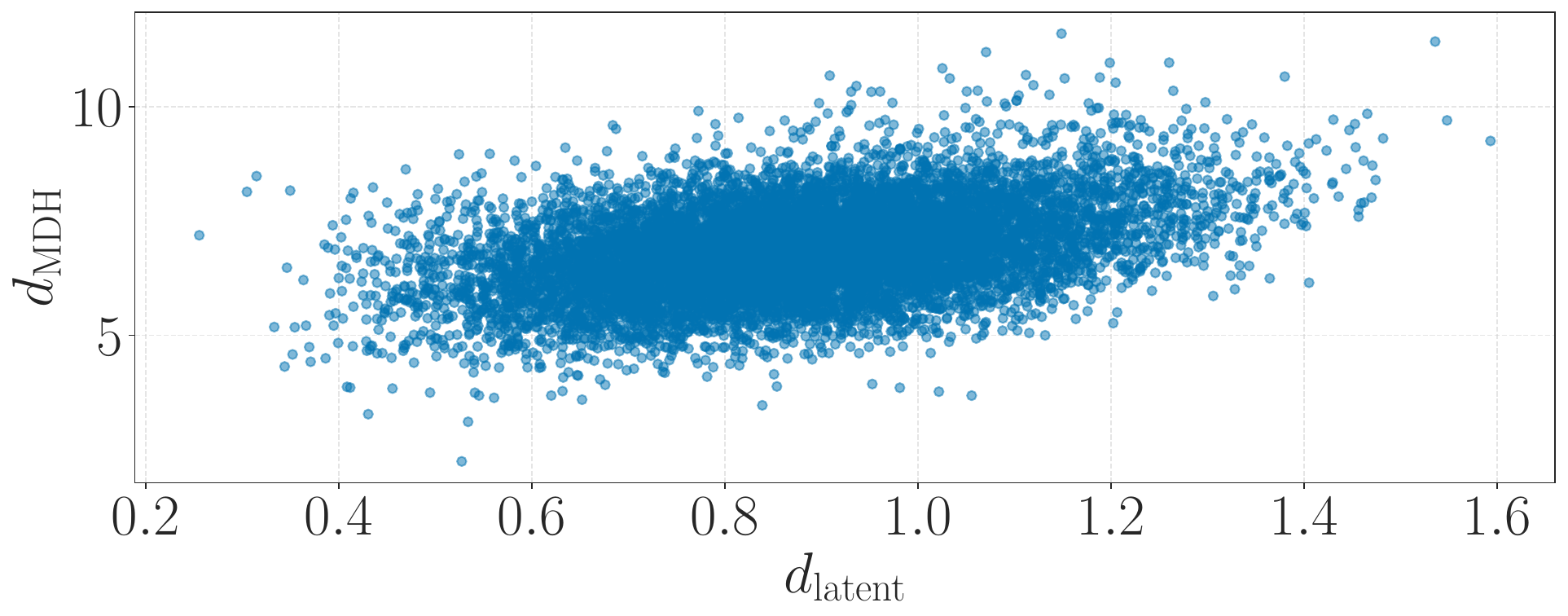}}\hfill
\subcaptionbox{$d_\text{workspace}$ vs. $d_\text{latent}$}[0.5\linewidth]{\centering\includegraphics[scale=0.2]{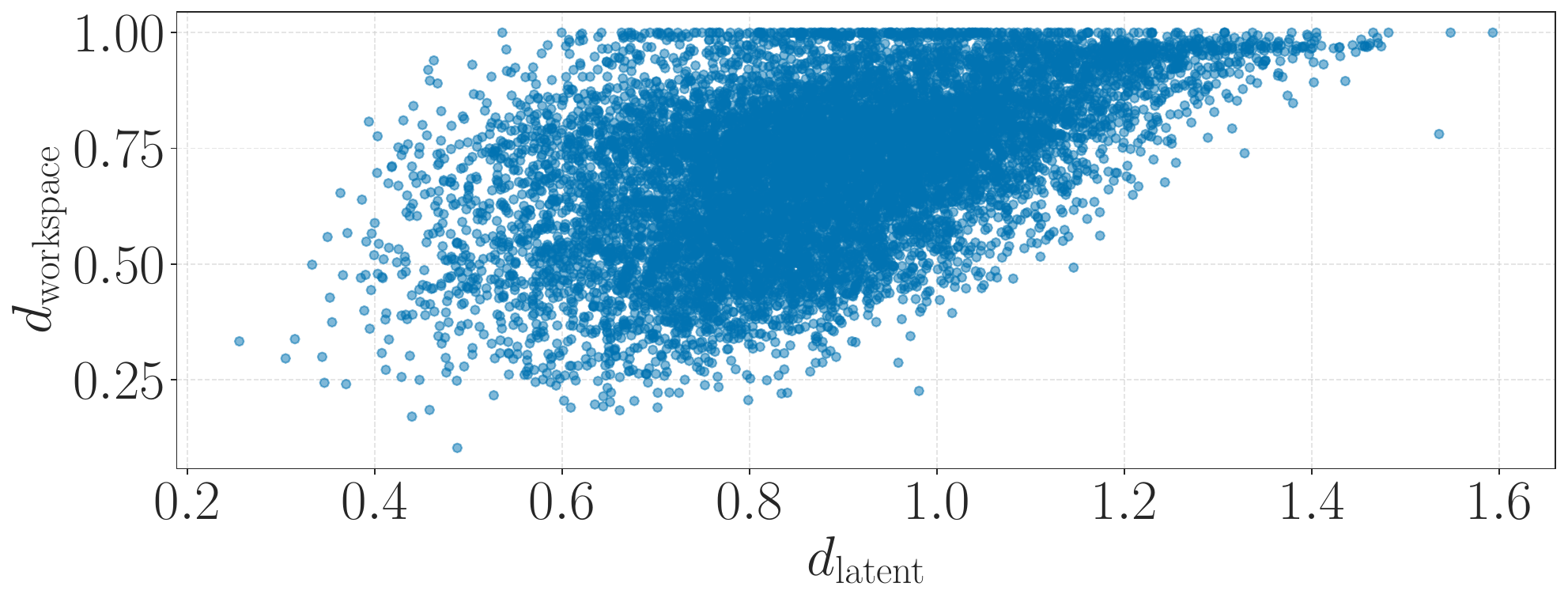}}\hfill

 \caption{Scatter plots of selected attributes against their most correlated principal component (a,b), and pairwise latent similarities against input and workspace similarities (d,c).}
 \label{fig:exploration}
\end{figure}

Second, we examined whether pairwise relationships in the latent space correspond to morphological and functional similarities. We sampled 150 morphologies and constructed $150\times150$ pairwise distance matrices in latent space, parameter space, and workspace. The parameter distance was computed as the Euclidean distance between flattened modified Denavit-Hartenberg parameters after scaling each parameter column by its sample standard deviation. Workspace distance was estimated using a Monte Carlo approximation of the Jaccard index~\citep{JaccardIoU1901}: for each morphology, $10^3$ reachable poses were sampled according to \Cref{ssec:pose_sampling}, and each pair was evaluated on the combined set of $2\cdot10^3$ poses using inverse kinematics. Mantel tests~\citep{Mantel1967} were then applied to the upper-triangular entries of the corresponding matrices, with significance assessed by row/column permutations and 95\% confidence intervals obtained by bootstrap resampling. \Cref{fig:exploration} shows that latent distance correlated more strongly with workspace distance at 49\% than with parameter distance at 38\%, both statistically significant. The encoder, therefore, does not merely re-express the kinematic parameters, which would preserve parameter distances most faithfully, but partially transforms them into a representation organised by the resulting workspace. A perfect correlation with workspace distance would mean that distances between embeddings fully mirror differences between workspaces. The remaining gap shows that the encoder organises the representation by workspace only in part, and that the decoder, which additionally receives the pose, completes the mapping to reachability.

\begin{table}
    \centering
    \caption{Spearman correlation coefficients (SCC) between morphological
             attributes and the most correlated principal component of the
             latent space, and Mantel test correlations between pairwise
             similarity matrices.}
    \label{tab:exploration}
    \begin{tblr}{colspec={l r r}, row{1}={font=\bfseries}}
        \toprule
        Attribute & SCC (\%) & $p$-value \\
        \midrule
        DoF $n$ & \num{87(1:0)} & $<$0.001 \\
        Moveable Length $r_\text{mov}$ & \num{30(3:3)} & $<$0.001 \\
        Base Twist $\alpha_0$ & \num{16(3:4)} & $<$0.001 \\
        Base Length $a_0$ & \num{61(2:2)} & $<$0.001 \\
        Base Offset $d_0$ & \num{27(3:3)} & $<$0.001 \\
        First Twist $\alpha_1$ & \num{7(4:3)} & $<$0.001 \\
        First Length $a_1$ & \num{11(4:3)} & $<$0.001 \\
        First Offset $d_1$ & \num{11(4:3)} & $<$0.001 \\
        EEF Twist $\alpha_n$ & \num{43(3:3)} & $<$0.001 \\
        EEF Length $a_n$ & \num{32(3:3)} & $<$0.001 \\
        EEF Offset $d_n$ & \num{53(2:3)} & $<$0.001 \\
        Reachability $\frac{\sum^{1000}_il_i}{1000}$ & \num{65(2:2)} & $<$0.001 \\
        Centre X $\left(\bm{t}_0\right)_x$ & \num{61(2:2)} & $<$0.001 \\
        Centre Y $\left(\bm{t}_0\right)_y$ & \num{42(3:3)} & $<$0.001 \\
        Centre Z $\left(\bm{t}_0\right)_z$ & \num{29(4:3)} & $<$0.001 \\
        Centre Magnitude $\anynorm{\bm{t}_0}_2$ & \num{30(3:3)} & $<$0.001 \\
        Min. Link Length $\min_i\sqrt{a_i^2+d_i^2}$ & \num{19(4:3)} & $<$0.001 \\
        Max. Link Length $\max_i\sqrt{a_i^2+d_i^2}$ & \num{53(3:2)} & $<$0.001 \\
        Std. Link Length $\sigma\!\left(\sqrt{a_i^2+d_i^2}\right)$ & \num{64(2:2)} & $<$0.001 \\
        Fraction Type 0 $\frac{\sum_i\mathds{1}\left(j_i =0\right)}{n}$ & \num{22(3:4)} & $<$0.001 \\
        Fraction Type 1 $\frac{\sum_i\mathds{1}\left(j_i =1\right)}{n}$ & \num{33(3:3)} & $<$0.001 \\
        Fraction Type 2 $\frac{\sum_i\mathds{1}\left(j_i =2\right)}{n}$ & \num{31(3:3)} & $<$0.001 \\
        Fraction Type 3 $\frac{\sum_i\mathds{1}\left(j_i =3\right)}{n}$ & \num{38(3:3)} & $<$0.001 \\
        Max. Pos. Type 3 $\max_i\mathds{1}\left(j_i =3\right)$ & \num{26(4:3)} & $<$0.001 \\
        \midrule
        \SetCell[c=3]{l}\textit{Mantel test} & & \\
        $d_\text{MDH}$ vs. $d_\text{latent}$ & \num{38(10:8)} & $<$0.001 \\
        $d_\text{workspace}$ vs. $d_\text{latent}$ & \num{49(12:10)} & $<$0.001 \\
        \bottomrule
    \end{tblr}
\end{table}

\section{Hyperparameters}\label{app:hyperparameters}
\Cref{tab:hyper} displays the hyperparameters of our classifier and the training routine, which were the result of a hyperparameter sweep~\citep{Optuna} on a fraction of the training set.
\begin{table}[H]
    \centering
    \caption{Hyperparameters of the classifier and the training.}
    \label{tab:hyper}
    \begin{tblr}{
            colspec = {l r},
            row{1} = {guard, font=\bfseries} 
        }
        \toprule
        Parameter & Value \\
        \midrule
        LSTM latent embedding size & $128$ \\
        LSTM \# layers & $1$ \\
        MLP hidden layer size & $1792$ \\
        MLP \# layers & $8$ \\
        \midrule
        Epochs & $1$ \\
        Batch size & $1000$\\
        Optimiser & Adam~\citep{KingmaAdam2015} \\
        Learning rate & $3\cdot10^{-4}$ \\
        $\bm{\beta}$ & $\left(0.9, 0.999\right)$ \\
        \bottomrule
    \end{tblr}
\end{table}

\end{document}